\documentclass[sn-basic, referee]{sn-jnl}

\usepackage{graphicx}%
\usepackage{multirow}%
\usepackage{amsmath,amssymb,amsfonts}%
\usepackage{amsthm}%
\usepackage{mathrsfs}%
\usepackage[title]{appendix}%
\usepackage{xcolor}%
\usepackage{textcomp}%
\usepackage{manyfoot}%
\usepackage{booktabs}%
\usepackage{subcaption}
\usepackage{listings}%
\usepackage{comment}%
\usepackage{array}

\usepackage{cleveref}
\usepackage{mathtools}
\usepackage{tikz}
\usepackage{diagbox}

\theoremstyle{thmstyleone}%

\theoremstyle{thmstyletwo}%
\newtheorem{example}{Example}%
\newtheorem{remark}{Remark}%

\theoremstyle{thmstylethree}%
\newtheorem{proposition}{Proposition}
\newenvironment{proofof}[1]
{\smallskip\noindent{\textbf{Proof~of~#1.}}
\hspace{1pt}}{\hspace{-5pt}{\nobreak\nobreak\hfill\nobreak
$\square$\vspace{2pt}\par}\smallskip\goodbreak}

\numberwithin{equation}{section}

\newcommand{\Lip}{\mathbf{Lip}}

\newcommand{\R}{\mathbb{R}}

\newcommand{\diff}{\mathop{}\!\mathrm{d}}

\makeatletter
\newcommand{\doublewidetilde}[1]{{%
  \mathpalette\double@widetilde{#1}%
}}
\newcommand{\double@widetilde}[2]{%
  \sbox\z@{$\m@th#1\widetilde{#2}$}%
  \ht\z@=.9\ht\z@
  \widetilde{\box\z@}%
}
\makeatother

\begin{document}

\title[The flat Metric]{Computing the Distance between unbalanced Distributions- The flat Metric}

\author*[1]{\fnm{Henri} \sur{Schmidt}}\email{henri.schmidt@posteo.de}
\equalcont{These authors contributed equally to this work.}

\author[1]{\fnm{Christian} \sur{D\"ull}}\email{duell@math.uni-heidelberg.de}
\equalcont{These authors contributed equally to this work.}

\affil[1]{\orgdiv{Institute of  Mathematics}, \orgname{Heidelberg  University}, \orgaddress{\street{Im Neuenheimer Feld 205}, \postcode{69120}, \city{Heidelberg}, \country{Germany}}}

\keywords{flat norm, dual bounded Lipschitz distance, Fortet-Mourier distance, unbalanced optimal transport}

\abstract{
 We provide an implementation to compute the flat metric in any dimension. The flat metric, also called dual bounded Lipschitz distance, generalizes the well-known Wasserstein distance $W_1$ to the case that the distributions are of unequal total mass. {\color{black} Thus, our implementation  adapts very well to mass differences and uses them to distinguish between different distributions. } This is of particular interest for unbalanced optimal transport tasks and for the analysis of data distributions where the sample size is important or normalization is not possible. The core of the method is based on a neural network to determine an optimal test function realizing the distance between two given measures. 
 Special focus was put on achieving comparability of  pairwise computed distances from independently trained networks. We tested the quality of the output in several experiments where ground truth was available as well as with simulated data.
}

\maketitle

\section{Introduction}
\noindent 
This paper is devoted to  a method for computing the flat metric between two nonnegative Radon measures of potentially unequal total mass, realized by a neural network. Special focus lies on an implementation which allows for comparability of  pairwise computed distances from independently trained networks. To this end, we extend the Wasserstein framework developed by \citep{paper_nn_Lip} to the unbalanced case. \\
{\color{black}
The paper is structured as follows: In the remainder of the introduction we define the flat metric and give a short overview on unbalanced optimal transport. As will be evident from the definition, test functions for the flat metric have to be Lipschitz continuous so that we modify a neural network approach  for the Wasserstein metric \citep{paper_nn_Lip} to our setting.  Section 2 is devoted to the architecture of the neural network as well as the subsequent adjustment of the output via experiments with ground truth to compensate for systematic errors. In Section 3 we provide  experimental validation of our method and residual analysis, whereas the conclusion is given in Section 4. Additional information on the calibration of the method, the adaptive penalty as well as on the hyperparameters and experimental details can be found in the Appendix. Furthermore, it contains a novel analytical distance result for Dirac measures in the flat metric.
}\\
\noindent
\textbf{Background:} We will consider measures with different masses, so that we work in  
 $\mathcal{M}^+(\mathbb{R}^d)$, i.e. the cone of nonnegative, bounded real-valued Borel measures on $\mathbb{R}^d$. We equip $\mathcal{M}^+(\R^d)$ with the \textbf{flat metric} (or \textbf{dual bounded Lipschitz distance}, \textbf{Fortet-Mourier distance}) defined by
   \begin{align}
    \label{def:flat_metric}
       \rho_F(\mu,\nu)= \sup_{\|f\|_{BL} \leq 1}\int_{\R^d}  f  \diff(\mu-\nu). 
   \end{align}
The class of test functions is given by the bounded Lipschitz functions endowed with the norm $\|f\|_{BL} = \max\left(\|f\|_{\infty}, \, |f|_{\Lip}\right)$,
where $\|f\|_{\infty}=\underset{x\in \R^d}{\sup}\,|f(x)|$ and \mbox{$|f|_{\Lip}=\underset { x\neq y}{\sup}\, \frac {|f(x)-f(y)|}{|x-y|}$.}
Note that formulation \eqref{def:flat_metric} resembles the Kantorovich-Rubinstein duality  of the Wasserstein distance $W_1$,  i.e.  
    \begin{align}
    \label{def:wasserstein_1_distance}
        W_1(\mu,\nu)=\sup_{|f|_{\Lip}\leq 1} \int_{\R^d}f \diff (\mu-\nu).
    \end{align}
Coming from optimal transport (OT) theory \citep{cuturi2013sinkhorn,villani:2003,Villani}, the Wasserstein metrics define distances between probability measures which take into account the  geometry of the underlying state space. Consequently, distances with respect to the Wasserstein metrics are more informative than methods based on divergences \citep{grauman, ling, wasserstein_active_contours,villani:2003,Villani}. Note that the Wasserstein distances scale with the total mass of the measures $\mu,\nu$ and are thus not necessarily restricted to probability measures. However, by construction the distances are only applicable in conservative problems, i.e. only if $\mu(\R^d)=\nu(\R^d)$, as otherwise no optimal transport plan exists, see e.g. \citep[Remark 1.18]{ulikowska2013structured}.\\
\noindent In most applications the distributions are normalized to probability measures so that any initial mass difference between the distributions is usually irrelevant. However, if the data distributions can not be normalized, e.g. as the mass differences of the distributions are actually meaningful since the underlying process is not conservative, then the data has to be artificially renormalized for OT to be applicable or the OT approach has to be discarded. This problem naturally appears in population dynamics with growth and death processes, see for example \citep{SCHIEBINGER2019928, schiebinger_stationary} where the authors employed the (entropically regularized) Wasserstein metric to compute distances between single cell gene distributions of cell samples in order to infer developmental trajectories. To compensate for the inherent cell growth over time, the authors had to introduce an additional model function which eliminates the impact of increasing cell numbers.
However, these problems also occur in other areas, such as in imaging or seismic analysis where the signal intensities fluctuate or even oscillate around 0, so that data can not be normalized \citep{9152115,UOT_waveform_inversion}.\\
\noindent Thus, in recent years numerous approaches appeared to tackle these unbalanced OT tasks, see \citep{Chizat_Peyre,peyre_cuturi} for an overview of several approaches on unbalanced optimal transport. In contrast to classical OT,
there is no mass restriction with unbalanced OT so that mass between distributions can not only be transported, but also created or destroyed. Among the most important applications are generative adversarial networks \citep{balaji,Yang_Uhler}, domain adaptation \citep{unbalance_minibatchOT, unbalanced_CO_OT}, color transfer \citep{sonthalia2020dual} and outlier detection \citep{mukherjee21a,balaji}. Since the unbalanced OT schemes can choose to ignore parts of the distribution due to mass deletion, they are quite robust to outliers \citep{mukherjee21a,balaji}. \\
\noindent 
However, the aim of this work is not to find an optimal (unbalanced) transportation plan, but a reliable way to compare measures with each other via a reasonable metric which simplifies interpretability of the distances. From the purely theoretical side, the obvious candidate would be given by the well-established total variation (TV) norm 
    \begin{align*}
    \|\mu\|_{TV}:=\mu^+(\R^d)+\mu^-(\R^d),
    \end{align*}
where $\mu^+,\mu^-\in \mathcal{M}^+(\R^d)$ are the measures arising from Jordan decomposition theorem \citep[Theorem 3.4]{Folland.1984}. However, as $\|\cdot\|_{TV}$ completely ignores the underlying geometry, this norm is is not suited for data which is obvious when computing the TV distance between two Dirac measures of the form $$\delta_a(x)=\left\{\begin{array}{cc}1,& x=a\\ 0,& \text{else}\end{array}\right.$$ One readily computes the distance to be 
    \begin{align*}
    \|\delta_a-\delta_b\|_{TV}=\delta_a(\R^d)+\delta_b(\R^d)=2 \qquad \forall a,b\in \R^d, a\neq b,
    \end{align*}
independent from the distance of the support points $a,b$. So instead we choose the flat metric defined by \eqref{def:flat_metric}. Apart from convenient analytical properties, providing completeness and separability for the measure space $\mathcal{M}^+(\R^d)$ \citep{ MR3764634}, the flat metric acts as a suitable generalization of the 1-Wasserstein distance $W_1$ to unbalanced tasks, and is as such also geometrically faithful, at least locally (see \ref{eq:two_Diracs_same_mass}). This is illustrated by the following alternative characterization due to Piccoli and Rossi \citep[Theorem 13]{piccoli2014generalized} 
    \begin{align}
\label{piccolis_alternative_characterisation_flat_norm}
     \rho_F(\mu,\nu)= \inf_{\substack{\tilde \mu\leq \mu,\,\tilde\nu\leq \nu\\\|\tilde\mu\|_{TV}=\|\tilde\nu\|_{TV}}}\|\mu-\tilde \mu\|_{TV}+\|\nu-\tilde\nu\|_{TV}+W_1(\tilde \mu,\tilde\nu).
    \end{align}
The decomposition \eqref{piccolis_alternative_characterisation_flat_norm} of $\rho_F$ into terms with TV norm and the term with  Wasserstein distance admits the typical interpretation of mass transport vs. mass deletion: Any share $\delta \mu$  of the mass of $\mu$ can either be transported from $\mu$ to $\nu$ at cost  $W_1(\delta \mu,\delta \nu)$ or removed/generated at cost $\|\delta \mu\|_{TV}$. As such, the minimal "sub-measures" $\tilde \mu,\tilde \nu$ achieve an optimal compromise between the strategy of mass transportation and of removal/generation. With regard to the implementation of the flat metric, we expect both regimes to display different associated errors that have to be accounted for individually.\\
 The flat metric has been used in \citep{MR3534005} for inverse problems in imaging and recently to establish well-posedness theory for structured population models in measures on separable and complete metric spaces \citep{our_book_ACPJ}.\\
\noindent In view of \eqref{piccolis_alternative_characterisation_flat_norm} we note that the approach introduced in \citep{mukherjee21a} tends to come closest to our setting as they also introduced a TV norm constraint instead of the typical Kullback-Leibler divergence to introduce an unbalanced optimization problem. Nevertheless, we choose to compute $\rho_F$ via \eqref{def:flat_metric} and not \eqref{piccolis_alternative_characterisation_flat_norm}.\\
At this point, we would like to remark that the goal of our implementation is not to achieve superior computational performance over already established methods, but merely to provide another perspective.  Although our method can in principal handle distributions of arbitrary dimension,  the treatment of high-dimensional distributions generally requires more data points, so that our proposed method becomes computationally expensive for dimensions $d=20$ and higher. At this point we refer to a recent paper \citep{2024paper} which applies nonequispaced fast Fourier transform to speed up the computations for radial kernels in unbalanced optimal transport tasks, so that high-dimensional data sets can be handled efficiently.\\

\section{Methods}
\begin{figure}
     \centering
     \includegraphics[width=\linewidth]{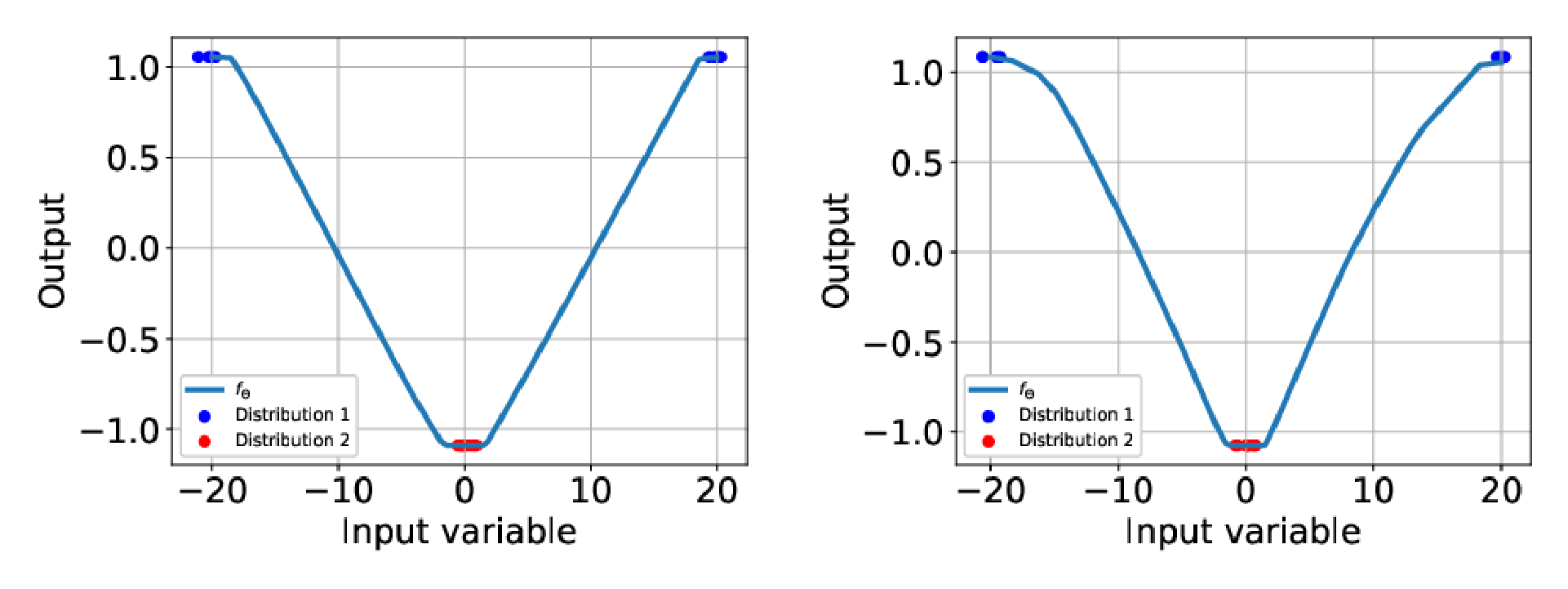}
     \caption{\small A simple 1D experiment showing the similarities between spectral normalization (left) and Bj\"orck orthonormalization (right). We considered two Gaussian mixture models $\mu=\frac{128}{2}(\mathcal{N}(-20, 0.5)+\mathcal{N}(20, 0.5))$ \textit{(blue)} and $\nu=128\mathcal{N}(0, 0.5)$  \textit{(red)}. In both cases the resulting $f_\Theta$ was plotted. Each time $f_{\theta}$ is bounded and $|f_{\Theta}|_{\Lip}\ll1$}
     \label{fig:bjorck_spectral}
\end{figure}

\noindent
Given two measures $\mu,\nu\in \mathcal{M}^+(\R^d)$, explicitly computing their flat distance via  \eqref{def:flat_metric} is highly nontrivial as finding a closed analytical expressions for the flat metric proves to be complicated even for Dirac measures, see \Cref{prop:general_formula}. So instead we trained a neural network of two fully connected hidden layers with 64 neurons each and the Adam optimizer \citep{KingmaB14}  to approximate $\rho_F(\mu,\nu)$ using \eqref{def:flat_metric}. Note that we deliberately chose a shallow network architecture as it provides sufficiently good results whereas moving to larger networks results in instabilities or even failures during training due to limited training data. In view of the Universal Approximation Theorem proven in \citep{paper_nn_Lip}, a suitable choice of architectural constraints allows the whole space $BL(\R^d)$ to be accessed via the network, so that we can expect meaningful results.\\
\noindent We make the ansatz $f=f_\Theta$ and model the optimal bounded Lipschitz test function by a multi-layer perceptron. To ensure that $f_\Theta$ is indeed admissible to the problem, i.e. that it is a bounded Lipschitz function with $\|\cdot\|_{BL}$ norm bounded by $1$, we use a mixed approach of regularization and architectural constraints. In particular, we adopt the architectural approach introduced in \citep{paper_nn_Lip} to guarantee Lipschitz continuity whereas we use regularizational constraints to account for the optimization problem \eqref{def:flat_metric} and to enforce boundedness of $f_{\Theta}$.\\
\textbf{Architectural constraints}:
In \citep{paper_nn_Lip} the authors Anil, Lucas and Grosse constructed a neural network to calculate the Wasserstein distance $W_1$ via its Kantorovich-Rubinstein duality \eqref{def:wasserstein_1_distance}.
 Their approach is based on the fact that  Lipschitz continuity is closed under compositions, so that it is sufficient to control the Lipschitz constant of each individual layer and activation function. 
In order to compute $W_1$ Anil, Lucas and Grosse proposed to normalize each layer $A_i$ and to use the 1-Lipschitz shuffling operator \textbf{GroupSort} \citep{chernodub2016norm} as activation function. This way the authors are able to construct a universal Lipschitz approximator. Hence, adopting the network architecture will yield Lipschitz continuity of $f_{\Theta}$.
We shortly summarize the most important concepts of the paper. \\
In \citep{paper_nn_Lip} the authors apply Bj\"orck orthonormalization \citep{Bjoerck}  during each forward pass which ensures that the linear transformation induced by layer $A_i$ is in fact isometric, thus strictly enforcing $|A_i|_{\Lip} = 1$. While this is convenient for the computation of $W_1$ as the test function $f$ will always be 1-Lipschitz theoretically, in our setting a Bj\"orck orthonormalization is too restrictive as in practice the optimal $f_\Theta$ of the flat distance often has a smaller Lipschitz constant $|f_\Theta|_{\Lip}$.
Thus, in our implementation we necessarily have to switch to \textbf{spectral normalization} $\|A_i\|_2=1$ instead which ensures that the largest singular value is $1$ but there may be other eigenspaces with smaller absolute singular values. In particular, we do not require $A_i$ to be 1-Lipschitz in every direction but just enforce $|A_i|_{\Lip} \leq 1$. As the spectral normalization- in contrast to Bj\"ork orthonomalization- is not gradient norm preserving, our choice potentially leads to  diminishing gradient norms of the network during backpropagation and thus to slower convergence of the network, see \cite[B.2]{paper_nn_Lip}.\\
\noindent
Nevertheless, the Bj\"orck and the spectral normalization yield similar results for a simple toy problem presented in \Cref{fig:bjorck_spectral}. In particular, the Bj\"orck approach is also able to produce gradients with norm less than one between probability measures $\mu, \nu$. This is rather surprising as in the Wasserstein case (i.e. without a bound constraint in the loss) $f_\Theta$ should indeed attain $|\nabla f_\Theta|=1$ due to the linear 1-Lipschitz layers, see also \cite[Corollary 1]{WGAN-GP}. We assume that the bound constraint \eqref{eq:loss_penalty} interferes with the normalization, such that the linear layers are in fact not completely orthonormal.

\begin{figure}
     \centering
     \includegraphics[width=\linewidth]{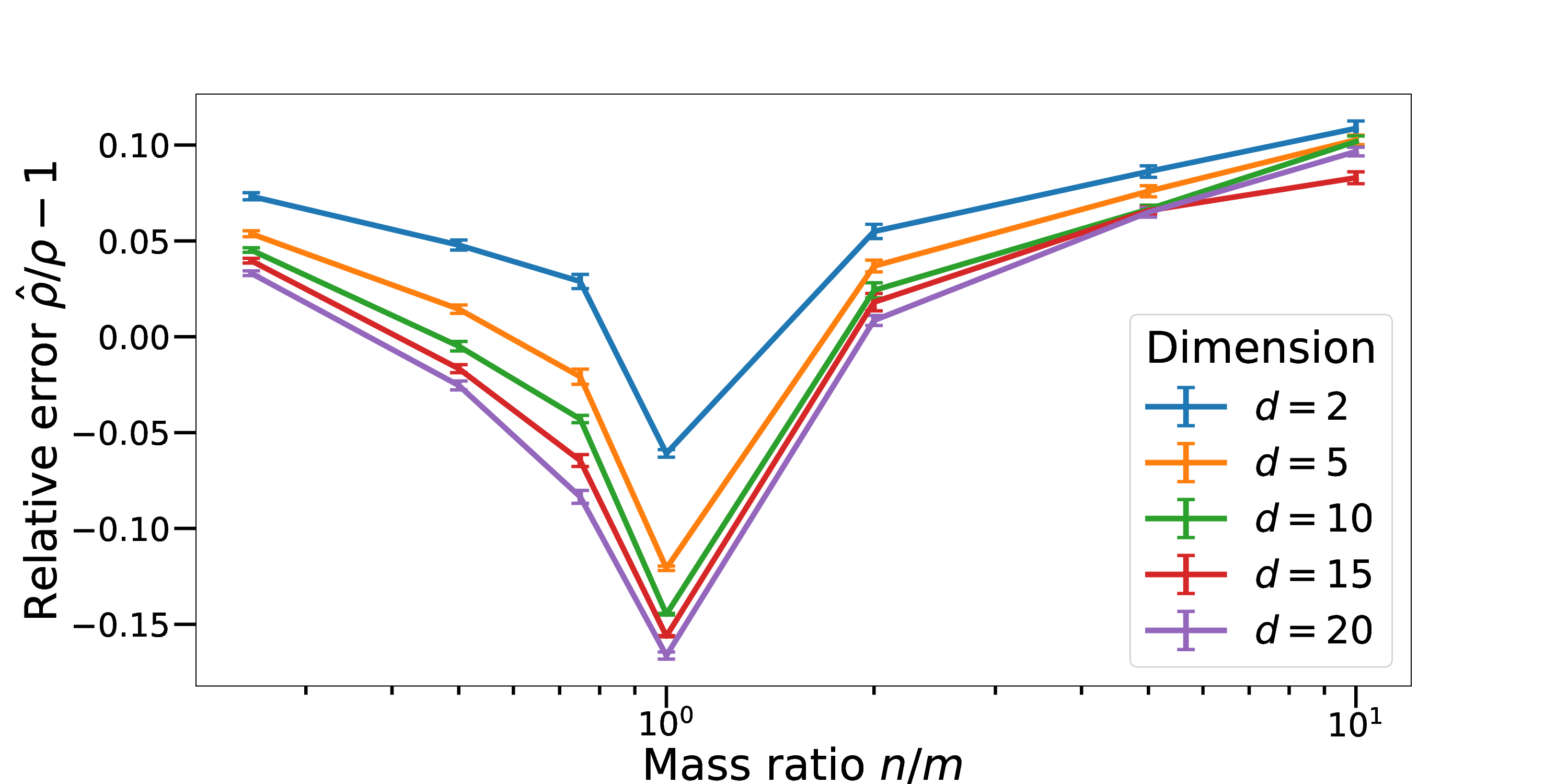}
     \caption{\small Relative error visualization of \Cref{relative_errors_averaged_over_radii}. Plotted are the incurred relative errors incurred in the calibration {\color{black} Experiment 1 } depending both on the mass ratio $n/m$ of $\nu$ and $\mu$ and the dimension. The resulting curve can be modelled by a negative log-normal distribution with a pronounced dip at equal masses ($n/m=1$). {\color{black}To improve the visualization and readability,  the x-axis uses a $\log$-scale.}}
     \label{fig:relative_error_curve}
\end{figure}

\noindent
The activation function \textbf{GroupSort} is a nonlinear, $1$-Lipschitz operator which generalizes \textit{ReLU} \citep{paper_nn_Lip}. It separates the pre-activations into groups and within each group permutes the input yielding an isometry. Typically, we will use two pre-activations per group, though higher values can be chosen too. In contrast to ReLU, GroupSort prevents gradient norm attenuation which would lead to $|f_\Theta|_{\Lip} \ll 1$ for deep networks. It often arises as a ReLU unit will map half of its input space to zero, thereby effacing all of the previous layers' gradients in this region. In fact, it can be shown  that a weight-constraint and norm-preserving neural network with ReLU activations is in fact linear, \citep{paper_nn_Lip}. Due to the lack of computational complexity, such a network is undesirable and thus the challenge is to construct a neural network which is 1-Lipschitz and simultaneously maintains enough expressive power to be a universal approximator. Both, in the work by Anil et al. \citep{paper_nn_Lip} and our work GroupSort has proven to work well while preserving enough expressive power to be a universal approximator.\\
\noindent Note that in view of \citep{NEURIPS2018} a Lipschitz constrained network provides provable adversarial robustness, i.e. the change in  output under small adversarial perturbations is bounded.\\
\noindent
\textbf{Regularization constraints}: Our loss term has to account for both the optimization problem of the flat metric and the boundedness constraint for $f_{\Theta}$, so that the \textbf{total loss term} $\mathcal{L}$ consists of two parts 
\begin{align}
\label{eq:total_loss_term}
    \mathcal{L} := \mathcal{L}_m + \lambda \mathcal{L}_b. 
\end{align}
The \textbf{metric loss} term $\mathcal{L}_m$ corresponds to minimizing the negative of \eqref{def:flat_metric} and is given by
\begin{align}
    \label{eq:loss_metric}
    \mathcal{L}_m:=-
    \int_{\R^d}f_{\Theta}(x)\diff \mu(x)+\int_{\R^d}f_{\Theta}(x)\diff \nu(x).
\end{align}
Note that after training $\mathcal{L}_m$ is our estimator for (the negative value of) the flat distance $\rho_F(\mu,\nu)$.\\
\noindent The additional penalty term to bound $f_{\Theta}$ is provided by the \textbf{bound loss term}
\begin{align}
    \label{eq:loss_penalty}
    \mathcal{L}_{b} &  \left(\frac{1}{\|\mu\|_{TV}}\langle h_\mu, h_\mu\rangle + \frac{1}{\|\nu\|_{TV}} \langle h_\nu, h_\nu\rangle\right),
\end{align}
where  $h_\kappa \coloneqq \max_{x\sim\kappa}(|f_\Theta(x)| - M, 0)$ and the parameter $M$ refers to the upper bound for $\|f_\Theta\|_\infty$ which in our formulation is given by $M=1$. By choosing this approach over simply considering the maximal value $\|f_\Theta\|_\infty$, we reduce the effect of outliers in the data, thus simplifying training. 
The auxiliary functions $h_\kappa$ encode in which areas $f_\Theta$ deviates from its target bound  evaluated each on the input given by $\kappa=\mu$ and $\kappa=\nu$ respectively. If such a deflection $|f_\Theta| > 1=M$ occurs, the corresponding $h_\kappa$ will have non-vanishing values in the appropriate domain and $h_\kappa$ serves as a penalty. The penalties are then accumulated over the whole space by the inner product $\langle\cdot,\cdot\rangle$, which thus  measures how much $f_\Theta$ violates the bound when evaluated with respect to $\mu$ and $\nu$ respectively. As the loss term should not favour measures with large total masses, we normalize each contribution by its respective total variation ensuring that the penalty terms remain invariant under scaling of the total mass. This will be useful as our implementation only considers discrete measures where the total variation is simply the number of support points so that it doesn't matter whether the same empiric distribution is given $100$ or $1000$ data points. \\
\noindent
The two penalty contributions with respect to $\mu$ and $\nu$ are then combined to give the overall penalty $\mathcal{L}_b$ incurred by violating bound $M$. In practice, enforcing the ideal bound of a vanishing $\mathcal{L}_b$ is not possible in general and hence we strive for small values of the loss. Due to the inner product, penalty contributions enter quadratically in $\mathcal{L}_b$ punishing larger deviations from $M$ more severely than smaller ones. \\
\noindent 
As 1-Lipschitz continuity of $f_{\Theta}$ will be guaranteed by the network architecture, the combined loss $\mathcal{L}$ then accounts for both rendering $f_\Theta$ admissible to the optimization problem \eqref{def:flat_metric} as well as finding the optimal value of the flat metric. 
Such an approach of having one loss term for the problem and one for the admissibility is commonly employed, e.g in the implementation of Wasserstein gradient-penalty adversarial networks \citep{WGAN-GP}. We remark that in \eqref{eq:total_loss_term} both contributions act antagonistically as a decrease in $\mathcal{L}_m$ often leads to an increase in $\mathcal{L}_b$, see  \Cref{fig:S_train_losses}, where the individual loss terms are monitored during  training. \\
\noindent 
Note that the two loss contributions $\mathcal{L}_m$ and $\mathcal{L}_b$ of $\mathcal{L}$ in \eqref{eq:total_loss_term} are effectively balanced by an enforcing parameter $\lambda=\lambda(t)$ which depends on the fraction of elapsed training $t$. Specifically, $\lambda$ is chosen \textit{adaptively} so that each freshly trained network is approximately bound by the same constant $\|f_\Theta\|_\infty \leq M$ while simultaneously having comparable relative loss contributions of $\mathcal{L}_m$ and $\mathcal{L}_b$  regardless of the input distributions. This is particularly important for our setting as we want to establish pairwise comparisons of neural networks which have been trained independently and/or on different data sets. 
This regularly occurs when computing pairwise distances between subdistributions so that the output of the network should be ordinal. Without proper balancing the resulting $f_{\Theta}$ will adhere more or less strict to the $\|\cdot\|_{\infty}$ bound depending on the currently dominating loss term leading to biased results. Notably, different networks would solve different optimization problems \eqref{def:flat_metric} yielding their actual outcomes to be incomparable to each other. Furthermore, each network requires a different optimal $\lambda$, so that we can not simply fix one sufficiently large value for $\lambda$  for the bound constraint $\|f_\Theta\|_\infty \leq 1=M$ to be satisfied in any case. Instead, we incorporated checks at various points during training, at which we update the enforcing parameter $\lambda$ dynamically to achieve comparable results.  Details to this procedure are listed in Appendix \ref{appendix:adaptive_penalty}.\\
\noindent
\textbf{Adjusting the output}\label{adjust_output} As the bound loss cannot vanish entirely in our implementation, it is to be expected that the raw output of our method will only approximately equal the correct theoretical value of the flat distance between the two given measures. In addition, it largely depends on the support of the measures whether the mass is predominantly transported or rather removed/generated. The different strategies can additionally lead to over- or underestimations of the true distance, depending on which prevails. To compensate for such systematic errors in the computation, we run a series of experiments where analytical ground truth is available and adjust the output accordingly. As closed analytical formulas results are difficult to find, we are restricted to comparatively simple distributions, see \Cref{prop:general_formula}.\\
\noindent {\color{black} \textbf{Experiment 1}: Up} to some scaling, we compute the distance between a Dirac measure with total mass $m\in \mathbb{N}$ located at the origin, and a linear combination $\nu$ of $n\in \mathbb{N}$ Dirac deltas with unit mass located in points $x_i$ on the $d$-dimensional hypersphere $S_{r_0}^{d-1}$ with radius $r_0$ around the origin, i.e.
	\begin{align*}
	\mu=m\delta_0,\qquad\qquad \nu=\sum_{i=1}^n\delta_{x_i}.
	\end{align*}
 We then vary the distance $r_0$ and average the resulting relative errors over the different radii to estimate the average error that would be expected in such a situation. The results are depicted in \Cref{relative_errors_averaged_over_radii} with more details in Appendix \ref{appendix:calibration}. We note that the relative errors are mostly of the same order of magnitude, which is a result of the adaptive penalty, see Appendix \ref{appendix:adaptive_penalty}. The visualization in \Cref{fig:relative_error_curve} suggests that the relative error follows a log normal distribution with a minor dependence on the dimension. The latter might also be a result of the fact that more data points are required as the dimension increases. Based on these findings, we correct the output of our implementation with a fitted log normal distribution that accounts for both the mass ratio of the measures involved and the influence of the dimension. \\
\begin{table}
    \begin{tabular}{c|ccccccc}
        \diagbox{dim}{$n/m$} & 0.25 & 0.5 & 0.75 & 1 & 2 & 5 & 10 \\\hline\vspace{-0.3cm}\\
	 2 &  0.073 & 0.048 & 0.0289 & -0.061 &  0.055 & 0.086 &  0.109 \\
        5 &  0.054 & 0.014 & -0.021 & -0.121 &  0.037 & 0.076 &  0.103  \\
        10 & 0.045 & -0.005 &-0.043 & -0.145 &  0.024 & 0.067 &  0.102  \\
        15 & 0.040 & -0.017 &-0.065 & -0.156 &  0.018 & 0.066 &  0.083   \\
        20 & 0.033 &-0.025 &-0.084  & -0.166 &  0.009 & 0.065 &  0.097 
    \end{tabular} 
    \caption{Relative errors of {\color{black} Experiment 1} in different dimensions and with varying mass ratios $n/m$ of the measures $\mu$ and $\nu$. For each parameter tuple $(dim,n/m)$ we randomly sampled support points of $\nu$ on spheres with prescribed radii $r_0\in\{0.5, 1, 2, 5\}$ and averaged the computed relative errors over $r_0$ (rounded to three decimals).}
 \label{relative_errors_averaged_over_radii}
\end{table}
\hspace{-0.2cm}{\color{black} \textbf{Experiment 2}}: In order to verify, whether the calibration proves to be effective, we conduct another test, where the output is corrected for the expected relative error. This time we drop the assumption that the support points of $\nu$ are located at prescribed radii of hyperspheres and instead allow for arbitrary support. According to the theory, it is more efficient to transport mass up to a distance of $2$, whereas beyond that range mass generation and deletion comes at a lower price. Hence, in our experiment we not only vary the mass ratio $n/m$ but also the fraction $l_f$ which denotes the share of $\nu$'s mass within the ball of radius $2$ around the origin. For more details we refer to \Cref{drop_support}. The results are enlisted in \Cref{tab:relative_errors_II}. It turns out that correcting the output can significantly reduce the effect of over- and underestimation, thus shrinking and homogenizing the relative errors. In comparison to the same experiment with uncorrected output, the mean of the absolute errors reduced by 34\% from 6.7\% (without correction) to 4.4\% (with correction).

{\color{black}\noindent
More specifically, we notice that the remaining residuals are caused by systematic and statistical effects to varying degrees. To better quantify this, each experiment -- i.e. each combination of $l_f$ and $n/m$ -- was repeated 50 times. \Cref{tab:relative_errors_II} details the resulting mean relative errors and their standard deviations. For some experiments, the remaining residual can be explained by stochastic variations. For instance, $l_f = 0.2$, $n/m = 2$ gives a residual of $\hat{\rho}/\rho - 1 = (1.0\pm2.4)\%$. As the standard deviation is significantly larger than the mean, this error can be well explained by some stochastic variation in the training progress. However, in roughly half the cases, this is not true since the standard deviation cannot explain the remaining residuals ($\text{std} > 3\,\text{mean}$). In such cases, we suspect systematic causes to play a role. Particularly, the largest deviations of around $\sim 10\%$ are found for intermediate $l_f \in\{0.2, 0.4, 0.6\}$, where both the moving and creation/deletion mode are vital. Most likely, these scenarios were not captured well enough during calibration, which mostly contains experiments of either pure transport or pure creation. Hence, a straightforward way to reduce systematic errors is to incorporate more test cases into the calibration setup. 
Similarly, the calibration process itself is subject to stochastic noise, which affects the fit of the log-normal distribution. Taking this into account, the method could be improved by not only using the best fit parameter, but rather employ error propagation such that each reported distance comes with its own uncertainty estimate.\\
\noindent
At a conceptual level, it may be that despite our efforts to achieve comparable effectiveness of the boundedness constraint via the Lagrange muliplier $\lambda \mathcal{L}_b$ in \eqref{eq:total_loss_term}, such comparability has not been sufficiently achieved leading to under- or overestimating of the distance in different experiments. Hence, incorporating a more sophisticated adaptation protocol for $\lambda$ should help with improving systematic deviations. Furthermore, it is possible that  correcting the output with the negative log-normal distribution is not the optimal way, so that a more elaborated approach could also lead to an improvement. 
}

\setlength{\tabcolsep}{2pt}
\begin{table}
    \centering
    \renewcommand{\arraystretch}{1.2}
    \begin{tabular}{
        c|*{6}{>{\raggedleft\arraybackslash}p{2.5em}@{\,}l}
    }
        \diagbox{$l_f$}{$n/m$} 
            & \multicolumn{2}{c}{0.5} 
            & \multicolumn{2}{c}{1} 
            & \multicolumn{2}{c}{2} 
            & \multicolumn{2}{c}{5} 
            & \multicolumn{2}{c}{10} 
            & \multicolumn{2}{c}{16} \\\hline
        0   & +1.3 & $\pm$ 1.0 & -2.8 & $\pm$ 1.2 & -6.7 & $\pm$ 1.5 & -7.9 & $\pm$ 1.7 & -6.4 & $\pm$ 1.3 & -5.2 & $\pm$ 1.2 \\
        0.2 & +14.4 & $\pm$ 1.3 & +6.8 & $\pm$ 1.9 & +1.0 & $\pm$ 2.4 & +1.0 & $\pm$ 2.9 & -0.9 & $\pm$ 1.7 & -4.7 & $\pm$ 1.5 \\
        0.4 & -2.2 & $\pm$ 1.5 & -6.3 & $\pm$ 1.8 & -13.3 & $\pm$ 2.5 & -15.0 & $\pm$ 3.2 & -9.7 & $\pm$ 4.2 & -3.0 & $\pm$ 1.1 \\
        0.6 & -0.8 & $\pm$ 1.1 & -13.2 & $\pm$ 1.2 & -7.6 & $\pm$ 1.3 & -4.4 & $\pm$ 0.9 & -2.7 & $\pm$ 0.7 & -0.1 & $\pm$ 0.4 \\
        0.8 & -1.6 & $\pm$ 1.3 & -4.0 & $\pm$ 1.2 & -1.9 & $\pm$ 0.9 & -1.3 & $\pm$ 0.6 & -0.6 & $\pm$ 0.4 & +0.1 & $\pm$ 0.2 \\
        1.0 & -2.7 & $\pm$ 1.3 & -2.3 & $\pm$ 1.1 & -1.9 & $\pm$ 0.8 & -1.9 & $\pm$ 0.5 & -1.4 & $\pm$ 0.3 & -0.7 & $\pm$ 0.2 \\
    \end{tabular}
    \caption{%
      Relative errors in percent ($\%$) of {\color{black} Experiment 2} in 2 dimensions with adjusted output according to the expected relative error (rounded to three decimals). Reported are the mean relative error and its standard deviation for $N=50$ repetitions of a cell's experiment. The fraction $n/m$ denotes the mass ratio of the measures $\mu$ and $\nu$. The parameter $l_f$ controls which fraction of the mass of $\nu$ is located within radius $2$ of the origin, i.e. the support point of $\mu$.
    }
    \label{tab:relative_errors_II}
\end{table}
\setlength{\tabcolsep}{6pt}
\noindent
\textbf{Implementation}
This paper and the corresponding code is based on the work by Anil, Lucas, and Grosse in \citep{paper_nn_Lip}. We forked their Github repository and adjusted it to our purposes. All our code can be found at \url{https://github.com/hs42/flat_metric} together with helpful beginner guides, examples and visualization tools.
\noindent
The code itself uses the PyTorch framework with unsupervised training. Notice that only the bound loss $\mathcal{L}_b$ acts as an error measure and should thus vanish after training whereas the metric loss $\mathcal{L}_m$ essentially becomes the estimator for the flat distance $\rho_F\approx - \mathcal{L}_m$ and hence ought to persist. \\
\noindent
The chosen network architecture of two fully connected hidden layers with 64 neurons each and the Adam optimizer \citep{KingmaB14} turned out to provide good results while moving to larger networks results in instabilities due to scarcity of training data for small distributions. In particular, the computed validation loss agrees well with the training loss and thus we conclude that our simple setup is powerful enough to generalize on the provided training set. This way, we can account for the inherent noise of experimental data and prevent overfitting. Concrete choices for hyperparameters as well as experiments on the performance depending on alternative network architectures can be found in \Cref{appendix:architectural_hyperparameter}.

\section{Experiments}
\subsection{High dimensional genomic data}
\begin{figure}
    \centering
    \includegraphics[width=0.7\textwidth]{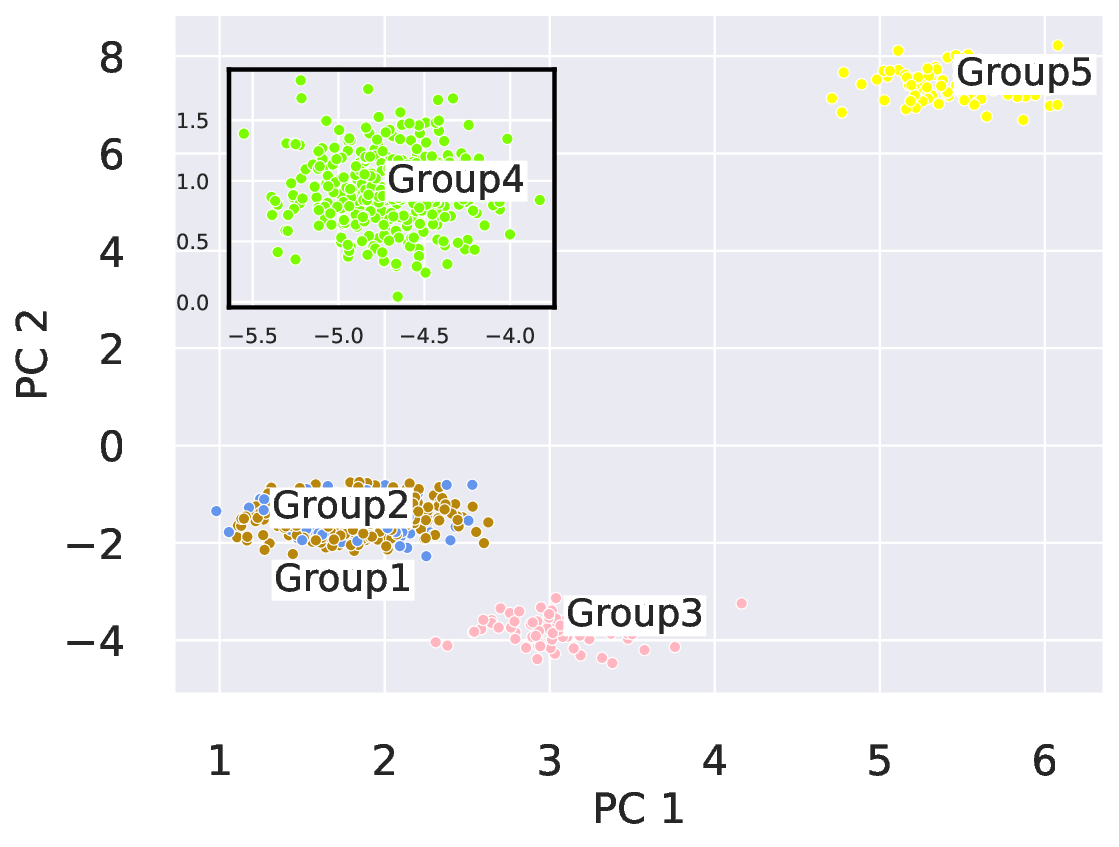}
    \caption{\small 2D PCA plot of mRNA counts for 5 distributions generated by \textit{Splatter}. Group1 (blue) and Group2 (brown) are nearly identical in the Wasserstein metric, but distinguishable in the flat metric case. Also note that in this plot Group4 is plotted in an inset to make for a better visualization.}
    \label{fig:tSNE_splatter}
\end{figure}

In order the test the implementation under more realistic circumstances, we conduct several experiments. In a first step, we analyzed  high dimensional simulated single-cell (sc) transcriptomics data generated by the \textit{R}-software package \textit{Splatter}. It was developed by Zappia et al. \citep{zappia2017splatter} to generate simulated scRNA sequencing count data of differentiation trajectories or of populations with one or multiple cell types. The simulation is based on a Gamma-Poisson distribution which models the expression levels of genes within cells as well as effects such as differing library sizes or dropouts. We refer to our Github repository for a  simulation script and a comprehensive workflow of the analysis. While there is no analytical ground truth available in this setting, we still have the possibility to monitor qualitative changes of the implementation via appropriate parameter choices in the \textit{Splatter} framework. In particular, we modelled five different cell groups by varying the sample size and the genetic expression profile, i.e. the location in gene space. A PCA-reduced visual is provided in \Cref{fig:tSNE_splatter}. After preprocessing and reducing the generated data to 5 dimensions, we determined the flat distances between the individual groups, see \Cref{tab:splatter}. For comparison, we compute the corresponding Wasserstein distances of the separately normalized distributions as well. 

\begin{table} 
        \begin{tabular*}{\textwidth}{@{\extracolsep\fill}lccccc}
        \midrule
        & Group 1 & Group 2 & Group 3 & Group 4 & Group 5\\

        Group 1 & (0.00, 0.00) & (4.27, 0.24) & (3.07, 7.21) & (4.75, 7.22) & (3.09, 9.94) \\ 
    
        Group 2 & (4.56, 0.25) & (0.00, 0.00) & (5.39, 7.23) & (2.89, 7.23) & (5.43, 9.95) \\

        Group 3 & (2.96, 7.19) & (5.05, 7.23) & (0.00, 0.00) & (5.06, 10.26) & (2.91, 12.14) \\
        Group 4 & (5.11, 7.17) & (2.89, 7.24) & (5.41, 10.25) & (0.00, 0.00) & (5.44,  11.99) \\
        Group 5 & (2.96, 9.94) & (5.09, 9.93) & (2.90, 12.18) & (5.10, 12.06) & (0.00, 0.00) \\
        \botrule
       \end{tabular*}
           \caption{\small Post-processed flat distances (first entry of each cell) between the clusters {\color{black} in 5 dimensions}. For comparison the respective Wasserstein distances using the same net architecture are displayed (second entry of each cell)}
             \label{tab:splatter} 
\end{table}
\noindent
One clearly notices the systematic differences between the flat metric and the Wasserstein distance.
As the latter is insensitive to population size, distributions $1$ (blue) and $2$ (brown) are nearly identical in Wasserstein space, whereas they are clearly distinguishable with respect to the flat metric due to the large mass difference. Taking the mass into account significantly influences the {\color{black} neigborhood} relation of the groups. {\color{black} In terms of Wasserstein distance, groups 1 and 2 are almost identical (distance 0.25), and the distance between group 1 and group 3 is extremely pronounced (7.19). In contrast, group 1 and 2 are clearly distinguishable in the flat distance (distance 4.56), so that group 3 is even the closest neighbor of group 1 (distance 2.96); on a par with group 5.} The same conclusions hold in a high-dimensional setting as well, see \Cref{fig:tSNE_splatter_dim50} and \Cref{tab:splatter_dim50}. Thus, if differences in cluster sizes are not only an effect of sampling but rather play a relevant role for the underlying question, we highly recommend using a method for unnormalized data distributions. Notice however, that the distances displayed in \Cref{tab:splatter}
- both with respect to the flat metric and the Wasserstein distance - are only ordinal and not cardinal.

\subsection{Domain adaptation}
\begin{figure}
	\centering
         \includegraphics[width=0.8\textwidth]{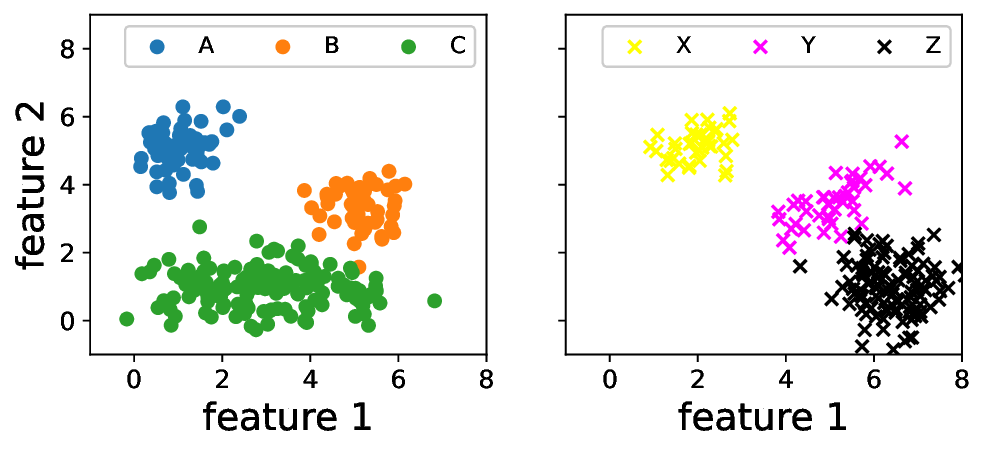}
     \caption{Example of domain adaptation. The three classes $A, B$ and $C$ in the source domain (left side) deviate in shape and mass from their targets $X, Y$ and $Z$ (right side). Distributions are modelled as multivariate Gaussians with the normalizations $m(A)=m(B)=1,\, m(C)=3,\, m(X)=0.75,\, m(Y)=0.85,\, m(Z)=2.25$}
     \label{fig:domains_adaptation}
\end{figure}

Domain adaptation refers to the task of identifying a learned data distribution in applied scenarios. The challenge consists in that the actually occuring samples show traits not necessarily covered in training, such that the target domain deviates from the known (and trained on) source domain.\\
\noindent
We now go on to show how unbalanced transport is naturally suited for such domain transfers as it allows for different volumes in feature space. In doing so, the flat metric offers a comparative advantage over similar implementations, e.g. \citep{mukherjee21a},  in that we are not constrained to some $\varepsilon$-imbalance in the distributions due to noise. Instead, the hallmark of the flat metric consists in its ability to handle systematic mass differences. This leads to a more natural matching of differently sized distributions of the target and source domain, since imbalances now act as an additional identifier during matching.  Consequently, simply normalizing the distributions to the same mass would discard information about the prelevance of the distributions. \\
\noindent
To illustrate, how the mass differences can  help to identify correct correspondences classes, consider the following example consisting of three classes, e.g. bicycle types. Their representations in the source space are known and labelled $A, B$ and $C$ as shown in \Cref{fig:domains_adaptation}. In real life, however, those bicycle types typically do not align perfectly with the learned distributions. Instead their target distributions $X, Y$ and $Z$ (originating from $A, B$ and $C$ respectively) deviate in shape and mass from their original sources, for instance due to difficulties during data acquisition, different measuring techniques and lost samples. 
\noindent
The distances between presented and learned classes are computed in the flat and the Wasserstein topology (cf. \Cref{tab:domain}).

\begin{table}

        \begin{tabular*}{\textwidth}{@{\extracolsep\fill}lccc}
        \midrule
        & Group A & Group B & Group C\\

        Group X & (1.16, 0.95) & (2.46, 3.62) & (5.13, 4.31) \\ 
    
        Group Y & (2.43, 4.35) & (0.42, 0.25) & (4.43, 3.32) \\

        Group Z	 & (3.35, 6.80) & (2.93, 2.73) & (2.06, 3.48)  \\
        \botrule
       \end{tabular*}
        \caption{\small Post-processed flat distances (first entry of each cell) between the clusters. For comparison the respective Wasserstein distances using the same net architecture are displayed (second entry of each cell)}
        \label{tab:domain}
\end{table}
\noindent
We observe that in the flat topology the matches are X: A, Y: B, and Z: C as given by the least distance. In the typical OT Wasserstein case, however, there is a mis-match as now the groupings read X: A, Y: B, Z: B. This is mainly due to the fact that both $Y$ and $Z$ are close to the source distribution $B$ in the Wasserstein space such that OT distances like the Wasserstein metric cannot distinguish between those. For UOT, however, the mass difference between $B$ and $Z$ discourages a match, and hence leads to the correct line-up.

\subsection{Residual analysis with benchmark datasets}
\begin{figure}
\centering
    \includegraphics[width=\textwidth]{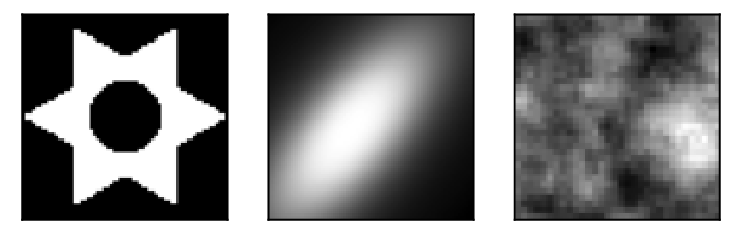}
        
\caption{Representatives of benchmarked image classes: geometrical shapes (left), bivariate Cauchy densities (middle) and Gaussian random fields (right).}
\label{fig:dotmark_figs}
\end{figure}

Lastly, we benchmark our implementation of the flat metric against the \textit{DOTmark} dataset. Devised by Schrieber et al \citep{dotmark} the \textit{Discrete Optimal Transport benchMARK} consists of ten different classes of grayscale images. Each class comprises different motives with resolutions varying from $32\times 32$ to $512\times 512$ pixels. It serves as a collection of problems to benchmark the performance of new OT techniques and validate their performance and has also been studied in unbalanced transport cases \citep{2024paper}.\\
\noindent
As the flat metric solves an unregularized optimization problem, comparable results by other teams are hard to find. Thus, we opted to analyze such cases within the DOTmark framework, where analytical ground truth is available. In light of \Cref{final_formula} we computed the flat distance between an image as the distribution $\mu$ and a single pixel representing the Delta distribution $\nu$. The pixels of the image matrix were assigned coordinates on the grid $[0,1]^2$ and their intensity were binned to integer values between $0$ and $255$. 
We investigated three categories: geometrical shapes, a bivariate Cauchy density with a random center and a varying scale ellipse, as well as a Gaussian random field; going from clear cut shapes to smeared out intensities and noise. \Cref{fig:dotmark_figs} exemplifies those classes. Within each class, we analyzed the ten images, both in resolution $32\times 32$ and $64\times 64$. The post-processed flat distances $\tilde{\rho}$ were then compared to their ground truths $\rho_F$ by the residual $\|\rho_F - \tilde{\rho}\| / \rho_F$. Even though the ground truths $\rho_F$ varied by a factor of ten in the benchmark tests (typically ranging between $3 ... 30$), our implementation remains faithful to those cases and typically deviates by only $4\%$ (overall error). {\color{black}
Specifically, the median residual for Cauchy densities as well as Gaussian random fields is about $0.04$, while it scored slightly worse for geometrical shapes with a median residual of $0.06$.} This benchmark acts as a proof of concept and confirms that the flat distance is suited to unbalanced optimal transport tasks.

\section{Conclusion}
\noindent
In this paper, we introduced an implementation of the flat metric $\rho_F$ for nonnegative Radon measures without a mass restriction. Particular focus was put on  comparability of  pairwise computed distances from independently trained networks. The combination of architectural (spectral normalization,
GroupSort activation function) and regularization constraints (bound penalty loss $\mathcal{L}_b$) turned out to be effective for estimating the flat distance as shown in several experiments. Throughout the tests, varying the hyperparameters -- both of the network architecture as well as of the analyzed problems -- did not yield qualitative discrepancies of the output indicating that the default setup of the net is robust. Choosing the enforcing parameter adaptively considerably shrunk the fluctuations in the relative errors guaranteeing that pairwise comparisons of distributions are possible. As originally the output was biased towards too high values, we adjusted the output with a negative log-normal distributions depending on the dimension and mass imbalance of the considered distributions. \\
\noindent On the basis of various experiments, we showed that our implementation of the flat metric can adapt very well to mass differences and use them to distinguish different distributions.

\newpage

\backmatter

\bmhead{Supplementary information}
\section*{Declarations}
\subsection*{Conflict of Interest}
The authors declare that they have no conflict of interest.
\subsection*{Funding}
This work is funded by the Deutsche Forschungsgemeinschaft (DFG, German Research Foundation) under Germany 's Excellence Strategy EXC 2181/1 - 390900948 (the Heidelberg STRUCTURES Excellence Cluster). \\
Christian Düll is supported by the European Research Council (ERC) under the European Union’s Horizon 2020 research and innovation programme (project PEPS, no. 101071786).\\
\subsection*{Ethics approval}
Not applicable
\subsection*{Consent for publication}
The authors of this article consent to its publication.
\subsection*{Availability of data and materials}
Not applicable
\subsection*{Code availability}
The code is available at \url{https://github.com/hs42/flat_metric}
\subsection*{Author contribution statement}
Both authors contributed equally to this research. The idea for this paper is based on the thesis of Henri Schmidt at Heidelberg University.

\begin{appendices}
\setcounter{figure}{5}
\setcounter{table}{4}
\renewcommand\thefigure{\arabic{figure}}  
\renewcommand\thetable{\arabic{table}} 
\section{Methods}\label{app:methods}
\subsection{Calibrating the implementation}\label{sec:exp}\label{appendix:calibration}
\begin{figure}
    \centering
    \includegraphics[width=\linewidth]{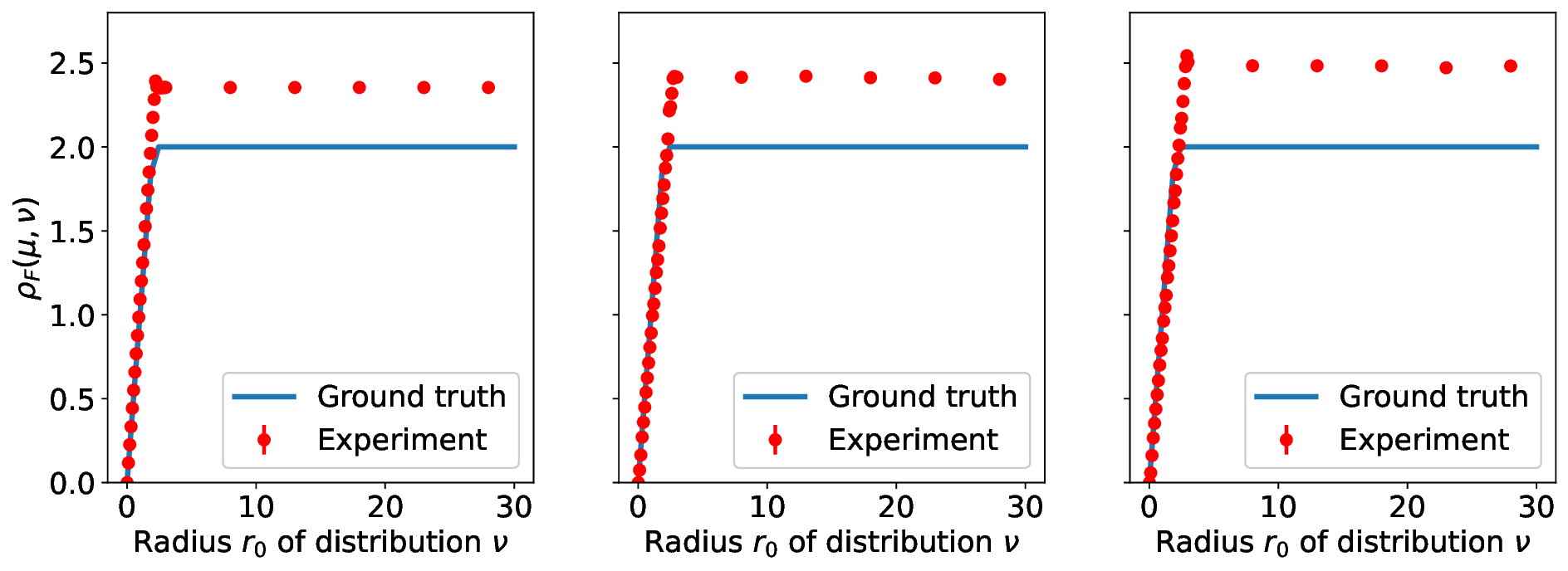}
    \caption{\small Estimates of $\rho_F(\mu,\nu)$ with balanced total mass (i.e. $n=m$) over varying radii $r_0\in\{0.01, 0.11, 0.21 \dots, 2.91, 3, 8, 13, 18, 23, 28\}$ and different dimensions $d$ (from left to right: $d=1$, $d=5$, $d=10$).  The flat distance is extracted as the mean of $-\mathcal{L}_m$ over the last $50$ training epochs with the corresponding error of the mean $\sigma/\sqrt{50}$ and adjusted according to the expected relative error.}
    \label{fig:saturating_r}
\end{figure}

As described in \Cref{adjust_output}, we expect our estimates to either under- or overestimate the true flat distance, according to whether mass is predominantly transported or removed and generated. To account for these systematic discrepancies, we performed {\color{black} the calibration Experiment 1} leading to \Cref{relative_errors_averaged_over_radii}, see the paragraph \textit{The calibration Experiment 1} for more details. As can be seen in  \Cref{fig:relative_error_curve} the relative errors as a function of the mass ratio between the distributions follows an inverted log-normal distribution. While we generally overestimate the ground truth, we significantly underestimate for balanced mass ratios. This behaviour holds true for all tested dimensions. To mitigate these tendencies, we fitted a function
\begin{equation*}
	f(x) = a\left(\exp\left(\frac{(x-\mu)^2}{2\sigma^2}\right) \frac{1}{\sqrt{2\pi}\sigma x}\right) + bx + c
\end{equation*} 
to the incurred relative errors $x=\hat{\rho}/\rho - 1$. The parameters $a,b,c,\mu,\sigma$ were thus determined for each of the considered dimensions $d=2,5,10,15,20$. We noticed that the parameters $p=(a,b,c,\mu,\sigma)^T$ depend linearly on the dimension, so that we chose the ansatz
\begin{equation*}
	p(d) = \alpha d + \beta
\end{equation*}
with $\alpha, \beta\in \mathbb{R}^5$. We proceeded as follows. Given the data in \Cref{relative_errors_averaged_over_radii}, we determined the optimal $\alpha$ and $\beta$ values to be
	\begin{align*}
	\alpha\approx 10^{-3}\times(-4.1, 0.1, -1.7, -2.0, 6.1)^T,\qquad \beta=10^{-1}\times(-1.1, 0.1, 0.5, 0.2, 0.2)^T,
	\end{align*}
 which in turn yields a dimension-dependent parametrization of the log-normal distribution. As the dimensionality of the problem and the mass ratio of the distributions are known à priori, we thus get an effective estimate on the the expected relative error $\hat{x}$ and can correct the output $\hat{\rho}$ accordingly: $\hat{\rho} \mapsto \hat{\rho} / (1 + \hat{x})$.\\
\textbf{The basic experimental design}: In order to better assess the quality of our implementation and the established correction, we consider some simple scenarios in which a mathematical ground truth is available for the flat distance. To this end, measure $\mu$ will always be a Dirac measure with mass $m$ concentrated at the origin, whereas $\nu$ is given by a linear combination of $n$ Diracs  concentrated at  points $x_i\in \R^d$. For simplicity, we invoke the scale invariance of the flat metric and scale both measures with the lower mass such that at least one measure is normalized, i.e. we consider
    \begin{align}
    \label{eq:general_form_of_mu_nu}
        \mu=\frac{m}{\min\{m,n\}}\delta_0 \qquad\text{ and } \qquad \nu=\frac{1}{\min\{m,n\}}\sum_{i=1}^n\delta_{x_i}.
    \end{align} 
\textbf{A first experiment}
 As a first step we assume that both measures have the same total mass, i.e. $n=m$ and construct measures $\mu, \nu$ as empiric measures by drawing $n$ changing random vectors  $x_1,...,x_n\in \R^d$ located on the $(d-1)$-sphere $S^{d-1}_{r_0}$ of radius $r_0$. To account for the increase in surface area we couple the number of sample points to the dimension $d$ by setting $n\coloneqq 30\cdot2^d$.  We then vary the distance between the supports $r_0$ to monitor the effects on the flat metric and repeat the experiment for several dimensions $d$. For each $r_0$ and each $d$ we train a new neural network for $2000$ epochs, where we chose two hidden layers of $128$ neurons each as an architecture. The results are depicted in  \Cref{fig:saturating_r}. In this setting, the flat norm can be analytically computed to be
\begin{align}
    \label{eq:two_Diracs_same_mass}
    \rho_F(\mu,\nu)= n\min\{2,r_0\},
\end{align}
independent from the dimension $d$. The proof of formula \eqref{eq:two_Diracs_same_mass} is a straight forward generalization of the result \cite[1.32]{our_book_ACPJ} and almost follows the same lines. In view of the alternative characterization \ref{piccolis_alternative_characterisation_flat_norm}, we see that theoretically up to the threshold $r_0=2$ it is more efficient to transport mass and  beyond that it becomes cheaper to delete mass and create it anew, a behaviour that can also be qualitatively found in \Cref{fig:saturating_r}.  The corrected estimates are very robust with respect to the considered dimensions. Starting at $r_0 \approx 2.4$ the distance $\rho_F$ is systematically overestimated which stems from the bound loss $\mathcal{L}_b$ not vanishing completely. However, the overestimation happens to the same extent for all $r_0\geq 2.4$ and is stable in every dimension.
\noindent

\begin{figure}
    \centering
    \includegraphics[width=0.8\textwidth]{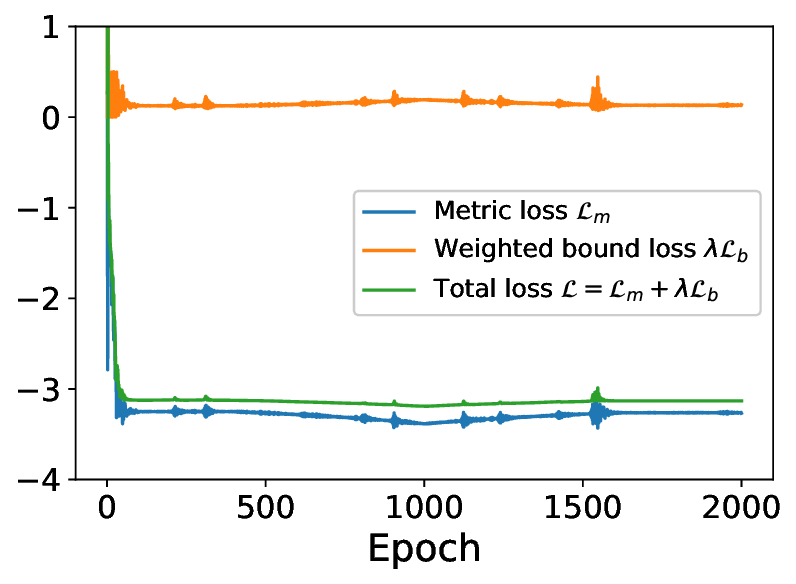}
    \caption{\small Loss terms during training of the experiment with $r_0=28$ and $d=2$}
    \label{fig:S_train_losses}
\end{figure}

\noindent
 In \Cref{fig:S_train_losses} we examined the mutual interactions of both loss terms. As can be seen both contributions act antagonistically as evident from the synchronous ripples and the dents around epoch 1000: Optimizing $\mathcal{L}_m$ first incurs at a cost on $\mathcal{L}_b$ and is then balanced by joint minimization of both constraints. 

\noindent \textbf{The calibration {\color{black} Experiment 1}} In a second step we allow for mass differences $n\neq m$ while still sampling the support points $x_i$ of $\nu$ on a sphere with prescribed radius $r_0$, i.e. $x_1,...,x_n\in S_{r_0}^{d-1}$. We then repeat the experiment for several combinations of mass ratios $n/m$ of $\mu$ and $\nu$ as well as different dimensions to monitor the effects. To account for the increase in surface area with the dimension we couple the number of sample points to the dimension by setting $n=2^d\tilde{n}, m=2^d$ with $\tilde{n}\in \{0.25, 0.5, 0.75, 1, 2, 5, 10, 30\}$. For each mass ratio $n/m$ and each dimension $d$ we train a new neural network for $10000$ epochs, where we chose two hidden layers of $128$ neurons each as an architecture. In this setting, the flat distance between $\mu$ and $\nu$ can analytically computed to be 
    \begin{align}
    \label{formula_simplest_experiment_2_diracs}
    \rho_F(\mu, \nu) = \min\{2, r_0\} + \frac{1}{\min\{m,n\}}|n-m|,
    \end{align}
which follows from adjusting the more general formula presented in \Cref{prop:general_formula}. As we have seen in the \Cref{fig:saturating_r} the behaviour of the relative errors depends on the distance of the supports, so that we average over the different radii to capture the expected relative error. The results are summarized in  \Cref{relative_errors_averaged_over_radii} and show that the relative error follows an inverted log normal distribution as a function of the mass ratio $n/m$. Furthermore, the dimension can be identified to have an influence, see \Cref{fig:relative_error_curve}. We then leverage these insights to correct the output of the implementation as discussed in the beginning of Appendix \ref{appendix:calibration}.\\
\noindent \textbf{Testing the effectiveness of the calibration {\color{black} with Experiment 2}}\label{drop_support} We test the effectiveness of the calibration with an extension of the previous experiment. In particular, instead of prescribing that the support poins $x_i$ of $\nu$ lie on fixed spheres with certain radii $r_0$, we randomly sample the $x_i$ within a ball with radius $200$ around the origin. In view of the fact that the flat distance theoretically becomes constant for mass transport beyond distance $2$, this should be sufficient to represent sampling in entire $\R^d$. As before, we scale both measures by the factor $1/\min\{m,n\}$ to guarantee that at least one measure is a probability measure. The measures $\mu$ and $\nu$ are thus again of the form \eqref{eq:general_form_of_mu_nu}. Without loss of generality, the $x_i$ are ordered with increasing distance to the origin, i.e. $d_1\leq d_2\leq...\leq d_n$ and let $l_f\in \{0,...,n\}$ be such that 
    \begin{align*}
        \begin{cases}
            d_i\leq 2, & i\leq l_f\\
            d_i>2, & i>l
        \end{cases}.
    \end{align*}  
In view of \Cref{formula_simplest_experiment_2_diracs} the parameter $l_f$ denotes the part of $\nu$ for which mass transportation is theoretically more efficient. According to \Cref{prop:general_formula},  in this setting the flat distance between $\mu$ and $\nu$ is analytically given by 
\begin{equation}
    \rho_F(\mu,\nu)=\frac{1}{\min\{n,m\}}\Big(|m-l_f|+n-l_f+\sum_{i=1}^{\lfloor \lambda\rfloor}d_i+(\lambda-\lfloor \lambda\rfloor)d_{\lfloor \lambda\rfloor+1}\Big),
\end{equation}
where $\lambda=\min\{l_f, m\}$ and $\lfloor\cdot\rfloor$ denotes the usual floor function. 
\noindent
In the experiment we varied the total masses $n,m$ as well as the fraction $l_f/n$ which indicates the percentage of $\nu$'s mass for which transportation is theoretically the better strategy. For each parameter set $(n/m,l_f)$ we uniformly sampled $\mathcal{O}(10^2)$ points $x_1,...,x_n$ in an open ball $B_{200}(0)\subset \R^2$ under the restriction that $l_f$ of them actually reside within $\overline{B_{2}(0)}\subset \R^2$. To prevent our network to overfit, we reduced the linear layers to $64$ neurons in both hidden layers and increased the number of training epochs to $10000$.

\noindent

\begin{figure}
        \centering
        \includegraphics[width=0.4\textwidth]{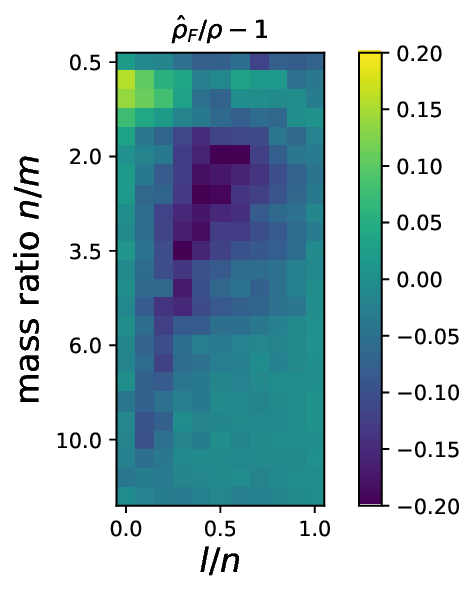}
        \caption{\small Relative errors of {\color{black} Experiment 2} for different parameter sets in dimension $d=2$. 
        The ``creation/deletion'' mode for the unbalanced optimal transport is attained for low $l/n$ (most data points outside ball of radius $2$) or high $n/m$ (highly differing total masses of the distributions), whereas actual transport is favoured for high $l/n$ (most data points inside ball of radius $2$) or $m/n\approx 1$. The visible diagonal separation marks a transition between both regimes}
        \label{fig:ratio_2_few_arbirary_m}
\end{figure}

\noindent
The results are summarized in \Cref{fig:ratio_2_few_arbirary_m}.  Apart from the upper triangular region bounded by $l/n\geq 0.1$ and $n/m \leq 6$, the errors are comparable indicating consistent quality of our implementation for distributions with highly unbalanced masses, where  mass creation/deletion is always the predominant mode. For higher ratios $l/n$ the network has to account for both mass transportation and mass creation which only works well in the confined region defined by the visible diagonal separation. Otherwise the  flat distance is underestimated in the upper region where $m\approx l$. In this case, $\mu$ and $\nu$ have roughly the same mass inside the ball $B_2(0)$ which is the region where mass transportation is more efficient. Theoretically, in this case all the mass inside $B_2(0)$ should be transported and all mass outside deleted. As it seems, the algorithm tends to underestimate mass transportation costs whereas it overestimates the cost of mass deletion. A repetition of the experiment in $d=4$ and $d=10$ dimensions yielded comparable qualitative behaviours. Postprocessing the output reduced the mean of the absolute errors each time by $21\%$ from $7.7\%$ (without correction) to $6.1\%$ (with correction) for $d=4$, and from $9.1\%$ to $7.2\%$ for $d=10$.

\subsection{The adaptive penalty}\label{subsec:unequal_masses}\label{appendix:adaptive_penalty}
\begin{figure} 
    \centering
    \includegraphics[width=0.6\linewidth]{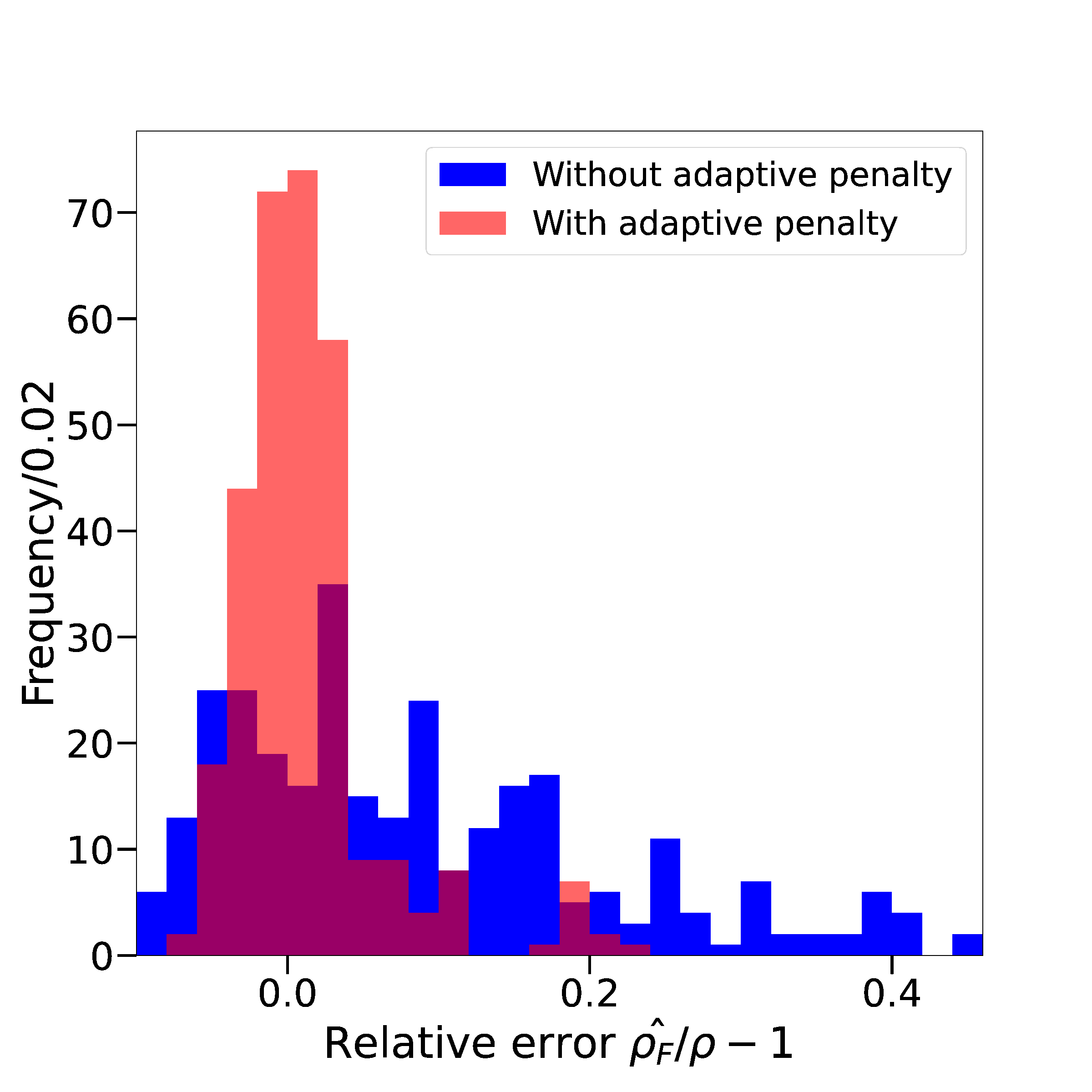}
    \caption{\small Histogram of the relative error distribution of corrected estimates with adaptive bound penalty (red) or not (blue), $d=2$, and bin width $0.02$. The non-adaptive bound penalty was fixed to $\lambda=10$}
\label{fig:histo_adaptive}
\end{figure}

\begin{figure}
    \centering
    \includegraphics[width=0.8\linewidth]{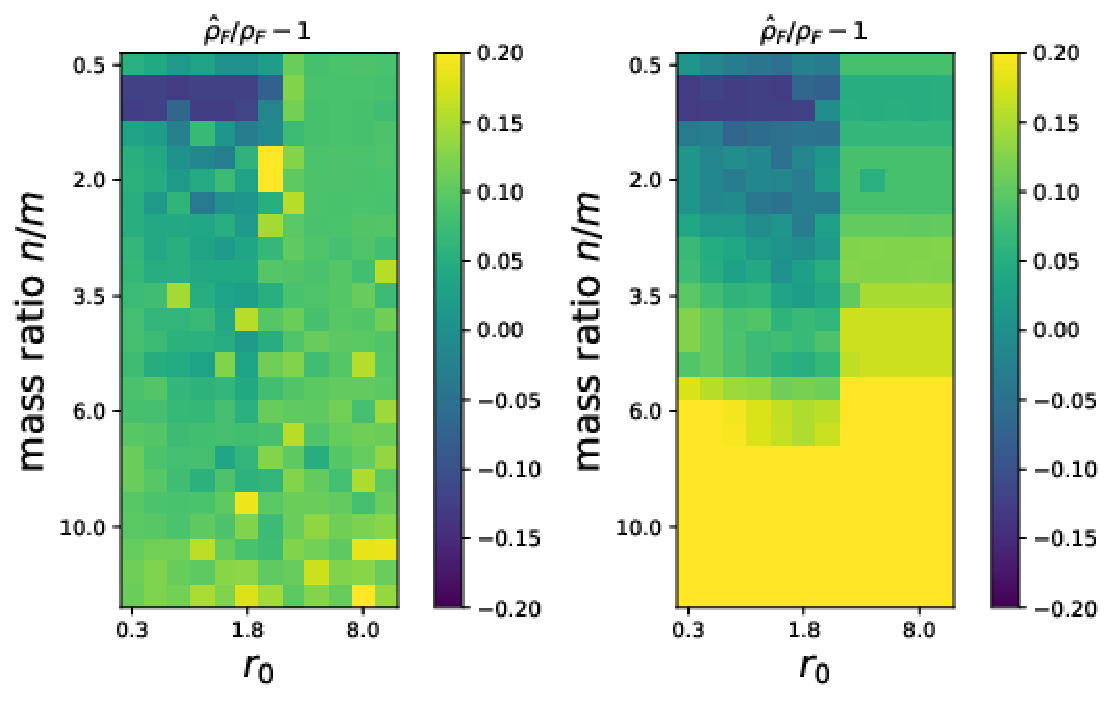}
    \caption{\small Relative errors  with (left) and without adaptive penalties (right) as a function of $r_0$ and the mass ratio $n/m$ of both distributions. The non-adaptive bound penalty was fixed to $\lambda=10$ and $d=2$ in both plots. }
\label{fig:ratio_2_few_with_and_without}
\end{figure}

One of the main goals of this paper is to provide an implementation of the flat metric which enables pairwise comparisons between several distributions. To this end, it has to be ensured that the outcomes of different networks are comparable to one another, i.e. that the same optimization problem is solved in all cases. To this end, we update the bound loss enforcing parameter $\lambda$ during runtime which guarantees that the different networks adhere to the bound loss constraint in the same way.\\
Notably, we start with an initial value $\lambda_{init}=10$ and then update at fractions $s_1=0.2$, $s_2=0.5$, and $s_3=0.8$ of all training epochs. The net is then trained until $s_1$ and we set $\lambda_{s_1} = - 2 \mathcal{L}_m$, i.e. to twice the current estimate for the flat distance. Thus, at $s_1$ we adapt the scale of $\mathcal{L}_b$ according to  $\mathcal{L}_m$. If the flat distance loss is high, so should be the penalty loss and both contributions are balanced to the same ratio. 
\noindent
After $s_2$, we instead look directly at the bound penalty $\mathcal{L}_b$ as we are interested in how much $f_\Theta$ exceeds the bound $M$. The actual penalty is compared to a target value of $b_t = 0.02$, i.e. we encourage each net $f_\Theta$ to disregard the boundedness constraint to the same extent by setting $\lambda$ to $\lambda_{s_2} = \lambda_{s_1}\frac{\mathcal{L}_b}{b_t}$. During the corresponding training epochs, $\lambda=\lambda(t)$ interpolates linearly between the target values $\lambda_{s_1},\lambda_{s_2}$ and $\lambda_{s_3}$. After $s_3$, $\lambda(t)$ remains constant. In summary, we have
\begin{equation*}
    \lambda(t) = \begin{cases}
                    \lambda_{init} & 0\leq t < s_1 \\
                    \frac{\lambda_{init} - \lambda_{s_1}}{s_2 - s_1}(t-s_2) + \lambda_{init} & s_1 \leq t < s_2 \\
                     \frac{\lambda_{s_1} - \lambda_{s_2}}{s_3 - s_2}(t-s_3) + \lambda_{s_1} & s_2 \leq t < s_3 \\
                     \lambda_{s_2} & s_3\leq t \leq 1
                \end{cases}.
\end{equation*}
To demonstrate the efficiency of the adaptive penalty, we repeated the experiment conducted to calibrate the implementation (see \textit{The calibration Experiment 1}) with a rigid scheme in two dimensions. The results are depicted in \Cref{fig:ratio_2_few_with_and_without} showing a significantly higher variance of the errors for different combinations of $n, m, r_0$. Without proper balance of the penalty terms, the error can hardly be controlled as soon as the mass ratio exceeds the critical threshold of $n/m\approx 6$. Furthermore, the histogram in \Cref{fig:histo_adaptive} shows the distribution of the relative errors of all considered parameter combinations either with or without an adaptive penalty. Adapting the penalty ensures that the distribution concentrates while simultaneously reducing the mean error from $ \hat{\rho}_F/\rho_F-1\approx 9.2\%$ (static case) to $\approx 1.5\%$ (adaptive penalty). A repetition of the above experiments with $6$ times more sample points and/or in dimension $d=4$ led to similar results.

\section{Hyperparameters}
\label{appendix:architectural_hyperparameter}
\begin{figure}
     \centering
    \includegraphics[width=\linewidth]{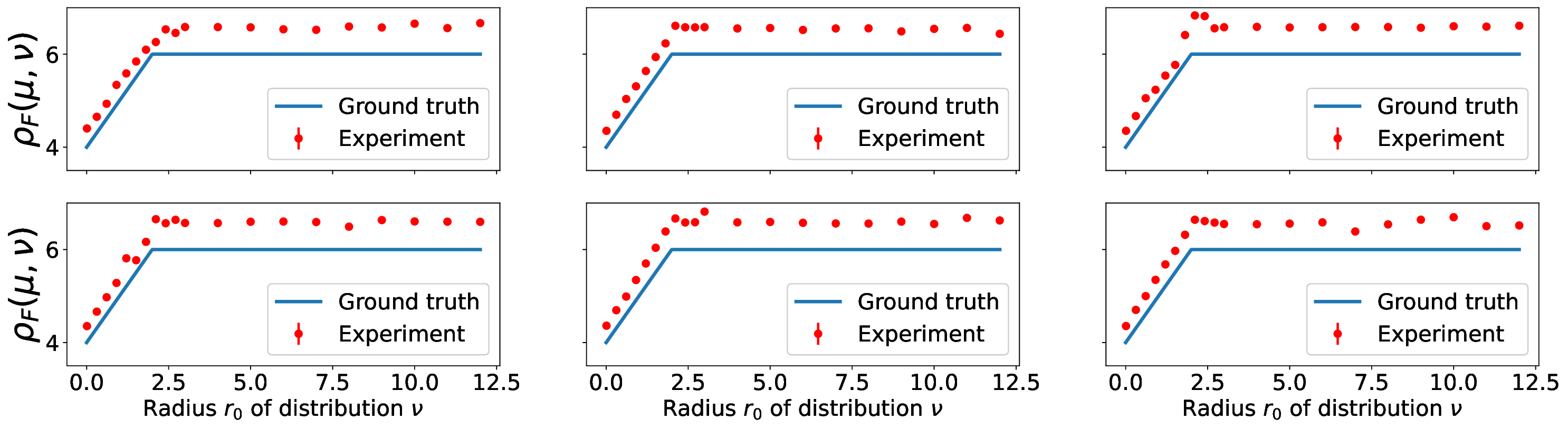}
    \caption{\small Showcases of different architectural hyperparameters. From left to right, top to bottom:
    SN with two hidden layers à $128$ neurons and bundles of $2$;
    BO with two hidden layers à $128$ neurons and bundles of $2$;
    SN with five hidden layers à $128$ neurons and bundles of $2$;
    SN with two hidden layers à $512$ neurons and bundles of $8$;
   BO with five hidden layers à $128$ neurons and bundles of $2$;
   BO with two hidden layers à $512$ neurons and bundles of $8$
   }
    \label{fig:S_hyperparas}
\end{figure}

Lastly, we examine the influence of the architectural hyperparameters on the performance of the neural network by repeating the experiment described in \Cref{subsec:unequal_masses}. Specifically, we considered the measures 
\begin{align*}
        \mu=\frac{m}{\min\{m,n\}}\delta_0 \qquad\text{ and } \qquad \nu=\frac{1}{\min\{m,n\}}\sum_{i=1}^n\delta_{x_i}
    \end{align*}
for points $x_1,...,x_n\in S_{r_0}^{d-1}$ with $n=2^4\times 50, m=2^4\times 10$ and $d=4$ and observed the qualitative change of the estimates with $r_0$. We both considered spectral normalization (SN) and Bj\"orck orthonormalization (BO) and changed the number of layers and the grouping size of Groupsort; each time $10000$ training epochs were used. The results are shown in \Cref{fig:S_hyperparas} where the top left depicts our control setup used in the other parts of this paper. While there are naturally quantitative differences, \Cref{fig:S_hyperparas} shows that there are no qualitative differences and no analyzed cases is obviously better suited to make predictions, i.e. is closer to the ground truth. A possible exception is that of Bj\"orck orthonormalization with large hidden layers and bundles of $8$ (bottom right), which shows some unexpected oscillations. We conclude that our usual architectural setup is suited for this specific and for similar tasks.
\noindent 

\noindent
Aside from these architectural hyperparameters we also probed those concerning the optimizer. These include learning rates, learning rate schedules, beta values for the Adam optimizer, and weight decays. Their default settings are hereby set to: a training length of $10000$ epochs, a learning rate of $0.01$, a learning rate decay of $0.9$ after milestone epochs $32$ and $64$ (step scheduler), an Adam optimizer with beta values $\beta=(0.9, 0.999)^T$, and no weight decay.
To illustrate their effects we investigated a $d=4$ dimensional setting, where $\mu$ is concentrated at the origin and $\nu$ is supported at a sphere of radius $r=5$. Both measures are set to carry the same normalization. \Cref{tab:hyperparas} lists different settings of the parameters and the resulting relative errors. One can observe only minute effect on the accuracy; in particular the default settings prove to be set to sensible values. 

\begin{table}
    \centering
    \begin{tabular}{l||llll}
        \diagbox{Tested feature\\(probed values)}{Incurred error} & & & & \\\hline\hline
        learning rates\\$(10^{-1},\,10^{-2},\,10^{-3},\,10^{-4})$ & -0.071 & -0.078 & -0.077 & -0.078 \\ \hline
        learning rate decay with a step scheduler,\\milestones at epochs 32 and 64 \\$(0.9,\,0.8,\,0.7)$ & -0.078 & -0.079 & -0.080 & \\\hline
        learning rate decay with a step scheduler,\\milestones at epochs 200 and 400 \\$(0.9,\,0.8,\,0.7)$ & -0.077 & -0.077 & -0.077 & \\\hline
        learning rate decay with a step scheduler,\\milestones at epochs 1000 and 2000 \\$(0.9,\,0.8,\,0.7)$ & -0.080 & -0.079 & -0.078 & \\\hline
        learning rate decay with an exponential scheduler\\$(0.95,\,0.9,\,0.85,\,0.8)$ & -0.045 & -0.038 & -0.080 & 0.270 \\\hline
        Adam beta values $(\beta_1, 0.999)$ \\ $\beta_1 = (0.95,\,0.9,\,0.85,\,0.8)$ & -0.078 & -0.078 & -0.077 & -0.077 \\\hline
        weight decay \\ $(0,\,0.01,\,0.05,\,0.1)$ & -0.077 & -0.096 & -0.073 & -0.076 \\
    \end{tabular}
      \caption{Relative errors incurred by different choices of hyperparameters. Save for the tested one, the parameters were set to their default values in each row. Only minute effects can be observed.}
      \label{tab:hyperparas}
\end{table}

\section{Experimental Details}
\subsection{Splatter}\label{app:splatter}

\noindent We first simulate the mRNA counts of a batch consisting $1000$ cells, each expressing $5000$ genes. These cells were divided into five distinct groups of expression profiles. By modifying the \textit{de.prob} parameter group three and four have been constructed to express more distinct genes compared to the reference expression profile and even more so for group five. On the other hand, group one and two were set to have similar expressed genes leading to an overlap in gene space. Furthermore, groups one, three, and five had $3.5$ times less cells than group two and four. Thus, we can observe the influence of the mass differences as well as the spread of the distributions on the flat metric. \\
\noindent 
The generated ``raw'' data was preprocessed in a standard way by filtering out highly variable genes, normalizing the data to the library size and centering. Lastly, a principal component analysis (PCA) selected the $5$ most important dimensions for the subsequent analysis. (cf. \Cref{fig:tSNE_splatter}).  
\noindent
To show that our solution works for arbitrary dimensions (save for the curse of dimensionality), we reduced the Splatter data also to $50$ instead of $5$ features and performed the same analysis. In doing so, we generated new data with the same parameters but for $10 000$ cells rather than $1000$. The conclusions are similar as for the low-dimensional case as can be seen from the results below:

\begin{figure}
\centering
\includegraphics[width=0.8\textwidth]{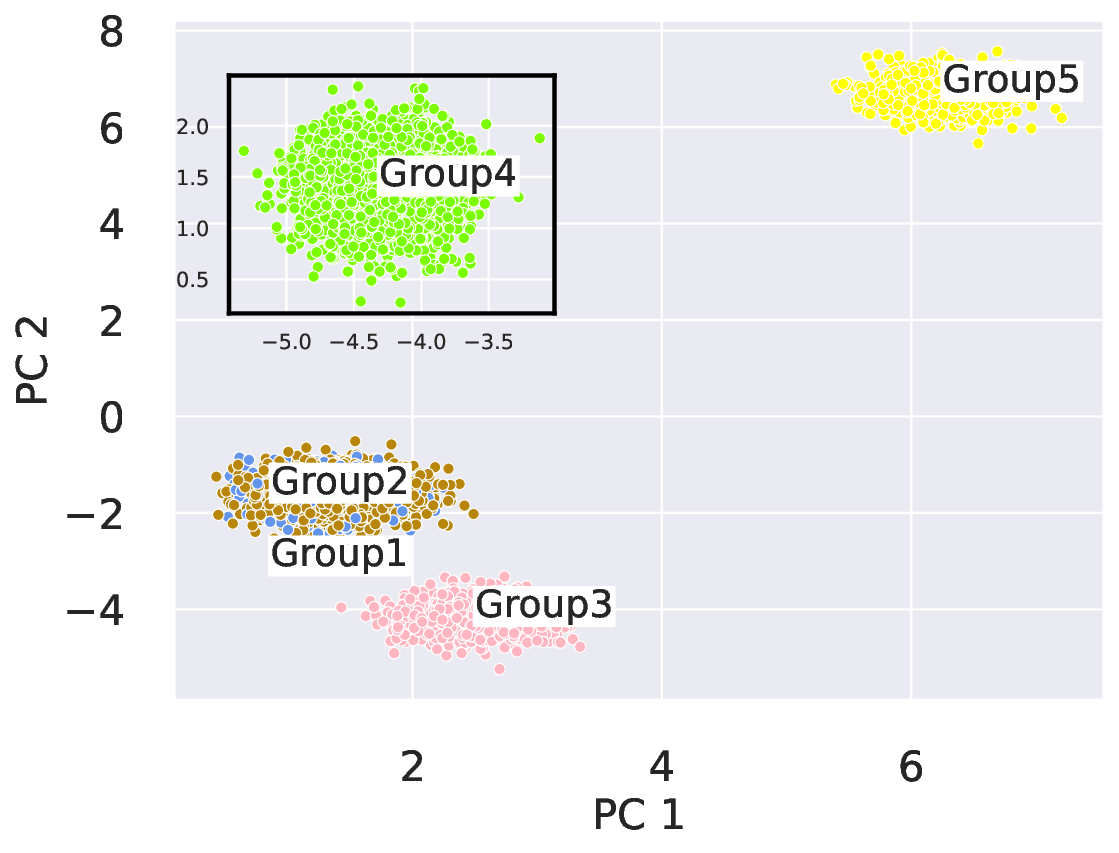}
\caption{\small 2D PCA plot of mRNA counts for 5 distributions generated by  \textit{Splatter}. The only difference to \Cref{fig:tSNE_splatter} is that population sizes were increased by a factor of $10$}
\label{fig:tSNE_splatter_dim50}
\end{figure}

\begin{table}
        \begin{tabular*}{\textwidth}{@{\extracolsep\fill}lccccc}
        \midrule
        & Group 1 & Group 2 & Group 3 & Group 4 & Group 5\\

        Group 1 & (0.00, 0.00) & (3.15, 0.35) & (2.89, 6.77) & (5.15, 6.85) & (2.93, 9.75) \\ 
    
        Group 2 & (3.38, 0.35) & (0.00, 0.00) & (5.50, 6.76) & (2.87,  6.85) & (5.18, 9.76) \\
        Group 3 & (2.88, 6.77) & (5.15, 6.76) & (0.00, 0.00) & (5.21, 9.51) & (2.94, 11.86) \\
        Group 4 & (5.49, 6.85) & (2.88, 6.68) & (5.54, 9.51) & (0.00, 0.00) & (5.21, 12.05) \\
        Group 5 & (3.04, 9.75) & (4.82, 9.76) & (3.07, 11.86) & (4.87, 12.05) & (0.00, 0.00)\\
        \botrule
       \end{tabular*}
        \caption{\small Corrected flat distances (first entry of each cell) between the {\color{black} cell} clusters for dimension $50$. For comparison the respective Wasserstein distances using the same net architecture are displayed (second entry of each cell)}
         \label{tab:splatter_dim50}
\end{table}

\subsection{Domain adaptation}
The classes $A,\,B,\,C,\,X,\,Y$ and $Z$ were modelled as bivariate Gaussians with means and covariances 
	\begin{align*}
\text{Cov}_A=\text{Cov}_B=&\sigma\begin{bmatrix}
1 & 0 \\
0 & 1\\
\end{bmatrix},\,\text{Cov}_C = \sigma\begin{bmatrix}
5 & 0 \\
0 & 1 \\
\end{bmatrix},\, \text{Cov}_X = \sigma \begin{bmatrix}
1 & 0.5 \\
0.5 & 1 \\
\end{bmatrix},\, \\\text{Cov}_Y =& \sigma \begin{bmatrix}
1.5 & 1 \\
1 & 1.5 \\
\end{bmatrix},\, \text{Cov}_Z = 2 \text{Cov}_A, 
\end{align*}
where $\sigma = 0.3$. Furthermore, 
	\begin{align*} 
\boldsymbol{\mu}_A = \begin{bmatrix}
1 & 5\\
\end{bmatrix}^T,\, \boldsymbol{\mu}_B =& \begin{bmatrix}
5 & 3.5\\
\end{bmatrix}^T,\, \boldsymbol{\mu}_C = \begin{bmatrix}
3 & 1\\
\end{bmatrix}^T,\, \boldsymbol{\mu}_X = \begin{bmatrix}
2 & 5\\
\end{bmatrix}^T,\, \\
\boldsymbol{\mu}_Y =& \boldsymbol{\mu}_B,\, \boldsymbol{\mu}_Z = \begin{bmatrix}
6.5 & 1\\
\end{bmatrix}^T.
\end{align*}
\noindent
The number of samples correspond to the normalizations and read \mbox{$N_A=N_B=50,\,N_C=150,\,N_X=38,N_Y=42,\,N_Z = 113$}; thus we imposed mass ratios relative to class $A$ of $m(A) = m(B)=1,\, m(C)=3,\, m(X)\approx 0.75,\,$\\$m(Y) \approx 0.85,\, m(Z)\approx 2.25$.

\subsection{Benchmark datasets}
We analyzed images of the categories: geometrical shapes, Cauchy density, and a Gaussian random field (namely the GRFmoderate class defined by $\sigma^2 = 1,\,\nu=1,\,\gamma=0.15$ according to \citep{dotmark}). The pixel coordinates in $[0,1]^2$ served as support points for the distribution of the image. The pixel intensity represents the amount of mass corresponding to the position of the pixel, i.e pixels with higher intensities were included more often in the data set to simulate this effect. To save on computational costs, the intensities were binned into either of the nine ranges $(0,28], \,(28,57],\,(57,85],$\\ $(85,113],\,(113,142],\,(142,170],\,(170,198],\,(198,227],\,(227,255]$ and each pixel was repeated according to its bin number. E.g. a pixel of intensity $23$ would count as black and be ignored, whereas a pixel of intensity $250$ would be fed eightfold into the training data set.
This image distribution $\mu$ was then compared to a single hot pixel at coordinates $(0,0)$ of intensity $c=\{100,1000\}$, i.e. $\nu=c\delta_0$.

\section{Analytical Ground Truth}
The following theoretical results are based on the work from the PhD thesis \cite[Section 2.4]{thesis_duell_2024}. They provide a closed analytical formula for the flat distance between a Dirac measure located at the origin and a linear combination of Diracs.
\begin{proposition}\label{prop:general_formula}
Let $n\in \mathbb{N}$ and $c\in \mathbb{R}^+$. Consider points $x_0,x_1,...,x_n\in \mathbb{R}^d$ which are ordered with increasing distance to $x_0$, i.e. for $d_i:=|x_0-x_i|$ we have \mbox{$d_1\leq d_2\leq \dots\leq d_n$} and let $l\in \{0,\dots,n\}$ be such that 
    \begin{align*}
    \begin{cases}
    d_i\leq 2,& i\leq l,\\
    d_i> 2, &i> l
    \end{cases}.
    \end{align*}
Furthermore, for $i=1,...,n$ consider weights $b_i\in \mathbb{R}^+$. 
Then the flat distance between the measures $\mu=c\delta_{x_0}$ and $\nu=\sum_{i=1}^nb_i\delta_{x_i}$ is given by
    \begin{align}
    \label{final_formula}
     \rho_F(\mu,\nu)=\sum_{i=1}^{I^*}b_id_i+\left(\min\left\{c,\sum_{i=1}^lb_i\right\}-\sum_{i=1}^{I^*}b_i\right)d_{I^*+1}+\left|c-\sum_{i=1}^lb_i\right|+\sum_{i=l+1}^nb_i,
    \end{align}
where 
    \begin{align*}
        I^*=\max\left\{I\in\{0,...,l\}\mid \sum_{i=1}^{I}b_i\leq c\right\}.
    \end{align*}
\end{proposition}

\begin{remark}
If the weights $b_i$ of the measure $\nu$ in \Cref{prop:general_formula} are all equal to $1$, i.e. $b_i=1$ for $i=1,...,n$, then formula \ref{final_formula} simplifies to
    \begin{align}
    \label{final_formula_simplified_weights}
     \rho_F(\mu,\nu)=|c-l|+n-l+\sum_{i=1}^{\lfloor \lambda\rfloor}d_i+(\lambda-\lfloor \lambda\rfloor)d_{\lfloor \lambda\rfloor+1},
    \end{align}
where  $\lambda=\min\{c,l\}$ and $\lfloor\cdot\rfloor$ denotes the usual floor function.
\end{remark}
\noindent To clarify formula \eqref{final_formula} we provide two examples.
\begin{example}
\begin{itemize}
\item Let ${\color{red}\mu=5\delta_0}$ and ${\color{blue}\nu=\sum_{i=1}^7\delta_{x_i}}$, where $d_i<2$ for $i=1,...,4$ and $d_i>2$ for $i=5,6,7$. See \Cref{pic:Dirac_norm_1}. Then formula \eqref{final_formula} reads
  \begin{align*}
         \rho_F(\mu,\nu)=&\sum_{i=1}^4d_i+\left(\min\left\{5,\sum_{i=1}^41\right\}-\sum_{i=1}^41\right)d_5+\left|5-\sum_{i=1}^41\right|+\sum_{i=5}^71\\
         =&\sum_{i=1}^4d_i+1+3=\sum_{i=1}^4d_i+4.
    \end{align*}
    \item Let ${\color{red}\mu=\frac 5 2\delta_0}$ and ${\color{blue}\nu=\sum_{i=1}^4\delta_{x_i}}$, where $d_i<2$ for $i=1,2,3$ and $d_4>2$. See \Cref{pic:Dirac_norm_2}. Then formula \eqref{final_formula} yields
    \begin{align*}
        \rho_F(\mu,\nu)=&\sum_{i=1}^2 d_i+\left(\min\left\{\frac 5 2,\sum_{i=1}^3 1\right\}-\sum_{i=1}^2 1\right)d_3+\left|\frac 5 2-\sum_{i=1}^31\right|+\sum_{i=4}^41\\
        =&\sum_{i=1}^2d_i+\left(\frac 5 2-2\right)d_3+ \frac 1 2+1= \sum_{i=1}^2d_i+\frac 1 2d_3+ \frac 3 2.
    \end{align*}
\end{itemize}
\end{example}

\begin{minipage}[b]{0.42\linewidth}
\begin{center}
     \begin{tikzpicture}[scale=0.8, font=\small]
        \draw[style=dashed] (0,0) -- node[below] {2} (2,0);
        \draw[style=dashed] (0,0) circle [radius=2cm];
        \fill[red] (0,0) node[label={$5\delta_0$}] {} circle(1pt);
        \fill[blue] (0.2,-1.5) node[label={3:$\delta_{x_3}$}] {} circle(1pt);
        \fill[blue] (-0.3,-0.4) node[label={180:$\delta_{x_1}$}] {} circle(1pt);
        \fill[blue] (0.7,0.7) node[label={right:$\delta_{x_2}$}] {} circle(1pt);
        \fill[blue] (2.3,0.3) node[label={$\delta_{x_5}$}] {} circle(1pt);
        \fill[blue] (-1.7,0.1) node[label={85:$\delta_{x_4}$}] {} circle(1pt);
        \fill[blue] (-2.6,1.4) node[label={$\delta_{x_7}$}] {} circle(1pt);
        \fill[blue] (-2.4,-1) node[label={270:$\delta_{x_6}$}] {} circle(1pt);
    \end{tikzpicture} 
    \captionof{figure}{\small $c=5$, $l=4$, $N=7$, $I^*=4$}
    \label{pic:Dirac_norm_1}
    \end{center}
\end{minipage}
\hspace{1ex}
\begin{minipage}[b]{0.49\linewidth}
\begin{center}
\begin{tikzpicture}[scale=0.8, font=\small]
        \draw[style=dashed] (0,0) -- node[below] {2} (2,0);
        \draw[style=dashed] (0,0) circle [radius=2cm];
        \fill[red] (0,0) node[label={$5/2\delta_0$}] {} circle(1pt);
        \fill[blue] (0.2,-1.5) node[label={3:$\delta_{x_3}$}] {} circle(1pt);
        \fill[blue] (-0.3,-0.4) node[label={180:$\delta_{x_1}$}] {} circle(1pt);
        \fill[blue] (0.7,0.7) node[label={$\delta_{x_2}$}] {} circle(1pt);
        \fill[blue] (2.3,0.2) node[label={$\delta_{x_4}$}] {} circle(1pt);
\end{tikzpicture}
\captionof{figure}{$c=5/2$, $l=3$, $N=4$, $I^*=2$}
 \label{pic:Dirac_norm_2}
 \end{center}
\end{minipage}

\begin{proofof}{Proposition \ref{prop:general_formula}}
We apply the alternative characterisation of Piccoli and Rossi \eqref{piccolis_alternative_characterisation_flat_norm} and consequently have to find the optimal submeasures $\tilde \mu, \tilde \nu$. As $\mu$ is a Dirac measure located at $x_0$, any submeasure of $\mu$ is of the form $\tilde \mu_{\alpha}=\alpha\delta_{x_0}$ for some $\alpha\in\left[0,\min\left\{c,\sum_{i=1}^nb_i\right\}\right]$. The parameter $\alpha$ denotes the share of the mass located at $x_0$ which we want to transport to $\nu$, and  $\alpha$ is thus à priori bounded by the minimum of the total masses of $\mu$ and $\nu$. We directly compute 
    \begin{align}
    \label{TV_distance_mu}
        \|\mu-\tilde\mu_{\alpha}\|_{TV}=(c-\alpha)\|\delta_{x_0}\|_{TV}=c-\alpha.
    \end{align}
As $\nu$ is a linear combination of Diracs, its submeasures $\tilde \nu$ are slightly more difficult. However, any $\tilde \nu$ is definitely of the form $\tilde \nu_{\alpha,\beta}=\sum_{i=1}^n\beta_i \delta_{x_i}$ with weights $\beta_i\in [0,b_i]$ satisfying $\sum_{i=1}^n\beta_i=\alpha$ as both submeasures need to have the same total mass. The weights $\beta_i$ indicate how much of the mass located at $x_i$ comes via transportation from $x_0$, whereas the rest $b_i-\beta_i$  has to be created. We compute
    \begin{align}
        \label{TV_distance_nu}
        \begin{split}
        \|\nu-\tilde\nu_{\alpha,\beta}\|_{TV}
        =\sum_{i=1}^n(b_i-\beta_i)\|\delta_{x_i}\|_{TV}
        =\sum_{i=1}^n(b_i-\beta_i)=\sum_{i=1}^nb_i-\alpha,
        \end{split}
    \end{align}
where we used that the total variation norm behaves linearly for nonnegative measures and that $\|\delta_x\|_{TV}=1$.\\
Next, we need to check the Wasserstein distance between $\tilde \mu_{\alpha}$ and $\tilde \nu_{\alpha,\beta}$. If $\alpha=0$, then clearly $W_1(\tilde \mu_{\alpha},\tilde \nu_{\alpha,\beta})=0$, so let $\alpha>0$ for the upcoming computation
    \begin{align}
        \label{Wasserstein_distance_not_optimized}
        \begin{split}
        W_1(\tilde \mu_{\alpha},\tilde \nu_{\alpha,\beta})=W_1\!\left(\alpha \delta_{x_0},\sum_{i=1}^n\beta_i \delta_{x_i}\right)= \int_{S}|x_0-y|\diff \!\left[\sum_{i=1}^n\beta_i\delta_{x_i}\right]\!(y)=\sum_{i=1}^n\beta_i|x_0-x_i|.
        \end{split}
    \end{align}
Here we used that the Wasserstein distance between a Dirac measure and an arbitrary probability measure $\eta$ is given by
    \begin{align*}
        W_1(\eta,\delta_{x_0})=\int_{\R^d}|x_0-y|\diff \eta(y).
    \end{align*}
In view of identity \eqref{piccolis_alternative_characterisation_flat_norm}, we combine \eqref{TV_distance_mu}, \eqref{TV_distance_nu} and \eqref{Wasserstein_distance_not_optimized}  to get the following estimate
    \begin{align}
    \label{total_bound_alpha_beta}
    \begin{split}
        \rho_F(\mu,\nu)\leq& \|\mu-\tilde \mu_{\alpha}\|_{TV}+\|\nu-\tilde \nu_{\alpha,\beta}\|_{TV}+W_1(\tilde \mu_{\alpha},\tilde \nu_{\alpha,\beta})\\
        =&c+\sum_{i=1}^nb_i-2\alpha+\sum_{i=1}^n\beta_id_i=: F(\alpha,\beta),
        \end{split}
    \end{align}
with $\alpha=\sum_{i=1}^n\beta_i$ so that the case $\alpha=0$ is included. As all possible submeasures are of the form $\tilde \mu_{\alpha}, \tilde \nu_{\alpha,\beta}$ we reduced the problem to minimizing $F$ subject to the constraint $\alpha=\sum_{i=1}^n\beta_i$. We claim that the global minimum is attained in $(\alpha^*,\beta^*)$
where 
    \begin{align}
        \label{optimal_choices_alpha_beta}
        \alpha^*=\min\left\{c,\sum_{i=1}^lb_i\right\} \qquad \text{ and }\qquad 
        \beta_i^*=\begin{cases}
         b_i,&i\leq I^*,\\
         \alpha^*-\sum_{i=1}^{I^*}b_i, &i=I^*+1,\\
         0,& \text{ else}
        \end{cases}
    \end{align}
and $I^*=\max\left\{I\in\{0,...,l\}\mid \sum_{i=1}^{I}b_i\leq c\right\}$.
In Remark \ref{rem:derive_optimal_choice} we give a  heuristic for this specific parameter choice. Before we prove the optimality, we show that \eqref{optimal_choices_alpha_beta} leads to \eqref{final_formula}. So we plug in $(\alpha^*,\beta^*)$ into $F$ and treat both cases of $\alpha^*$ separately.\\
\underline{Case 1: $\alpha^*= c$}\\
In this case $c\leq \sum_{i=1}^lb_i\leq \sum_{i=1}^nb_i$, so that 
    \begin{align*}
        F(\alpha^*,\beta^*)=&c+\sum_{i=1}^nb_i-2c+\sum_{i=1}^n\beta_i^*d_i
        =\sum_{i=1}^n b_i-c+\sum_{i=1}^{I^*}b_id_i+\left(c-\sum_{i=1}^{I^*}b_i\right)d_{I^*+1}\\
        =&\sum_{i=1}^{I^*}b_id_i+\left(c-\sum_{i=1}^{I^*}b_i\right)d_{I^*+1}+\left|c-\sum_{i=1}^l b_i\right|+\sum_{i=l+1}^nb_i
    \end{align*}
    as claimed.\newpage
\noindent  \underline{Case 2: $\alpha^*=\sum_{i=1}^{l}b_i$}\\
In this case we have $I^*=l$ so that
    \begin{align*}
        F(\alpha^*,\beta^*)=&c+\sum_{i=1}^nb_i-2\sum_{i=1}^lb_i+\sum_{i=1}^n\beta_i^*d_i\\
        =&c+\sum_{i=1}^nb_i-2\sum_{i=1}^lb_i+\sum_{i=1}^{I^*} b_id_i+\left(\alpha^*-\sum_{i=1}^lb_i\right)d_{I^*+1}\\
        =&\sum_{i=1}^{I^*} b_id_i+\left|c-\sum_{i=1}^lb_i\right|+\sum_{i=l+1}^nb_i
    \end{align*}
as desired.\\
We are left to prove that $(\alpha^*,\beta^*)$ yields the global minimum of $F$. To show this, we invoke variational inequality theory. First note that the domain of $F$ 
\begin{align*}
        X:=\left\{(\alpha,\beta)\in[0,c]\times \bigotimes_{i=1}^n[0,b_i]\mid \alpha - \sum_{i=1}^n\beta_i=0\right\}\subset \R^{n+1}
    \end{align*}
is nonempty, closed and convex. Furthermore,
$F:X\to \mathbb{R}$  is smooth, linear and thus convex. Let $x^*:=(\alpha^*,\beta^*)$. According to \cite[7.5]{geiger_kanzow_2002} the point $x^*$ is a global minimum of $F$ if $x^*$ solves the variational inequality
    \begin{align}
        \label{variational_inequality}
        VIP(X,\nabla F):=\nabla F(x^*)^T(x-x^*)\geq 0\qquad \forall x\in X.
    \end{align}
We compute for some $x=(\alpha,\beta)\in X$
    \begin{align*}
       &\nabla F(x^*)^T(x-x^*)=\begin{pmatrix}-2\\d_1\\\vdots\\d_n\end{pmatrix}^T\begin{pmatrix}\alpha-\alpha^*\\\beta_1-\beta_1^*\\\vdots\\\beta_n-\beta_n^*\end{pmatrix} 
       =-2(\alpha-\alpha^*)+\sum_{i=1}^n d_i(\beta_i-\beta_i^*)\\
       =&-2\left(\sum_{i=1}^n\beta_i-\sum_{i=1}^n\beta_i^*\right)+\sum_{i=1}^n d_i(\beta_i-\beta_i^*)=\sum_{i=1}^n (d_i-2)(\beta_i-\beta_i^*).
    \end{align*}
Plugging in $\beta^*$ yields
     \begin{align}
     \begin{split}
     \label{general_contraint}
       &\nabla F(x^*)^T(x-x^*)\\
       =&  \sum_{i=1}^{I^*}(d_i-2)(\beta_i-b_i)+(d_{I^*+1}-2)\left(\beta_{I^*+1}-\alpha^*+\sum_{i=1}^{I^*}b_i\right)+\sum_{i=I^*+2}^n(d_i-2)\beta_{i}.
       \end{split}
    \end{align}
To see that \eqref{general_contraint} is actually nonnegative for all $x\in X$ we have to distinguish cases for $\alpha^*$.\\
\underline{Case 1: $\alpha^*=\sum_{i=1}^lb_i$}\\
Then $I^*=l$ and \eqref{general_contraint} reads
    \begin{align*}
       &\nabla F(x^*)^T(x-x^*)\\
       =& \sum_{i=1}^{l}(d_i-2)(\beta_i-b_i)+(d_{l+1}-2)\left(\beta_{l+1}-\sum_{i=1}^lb_i+\sum_{i=1}^lb_i\right)+\sum_{i=l+2}^n(d_i-2)\beta_i\\
        =&\sum_{i=1}^{l}\underbrace{(d_i-2)}_{\leq 0}\underbrace{(\beta_i-b_i)}_{\leq 0}+\sum_{i=l+1}^n\underbrace{(d_i-2)}_{> 0}\beta_i\geq 0 \qquad \forall (\alpha,\beta)\in X.
       \end{align*}
In particular, $x^*$ solves the variational inequality \eqref{variational_inequality} and is thus the global minimum of $F$.\\
\underline{Case 2: $\alpha^*=c$}\\
In this case we have $I^*\leq l$ but we can assume without loss of generality that $I^*+1\leq l$ as otherwise $I^*=l$ and thus $c=\sum_{i=1}^lb_i$ which has already been covered in the first case. Hence, \eqref{general_contraint} reads
    \begin{align}
    \begin{split}
    \label{constraint_in_complicated_case}
         &\nabla F(x^*)^T(x-x^*)\\
         =& \sum_{i=1}^{I^*}(d_i-2)(\beta_i-d_i)+(d_{I^*+1}-2)\left(\beta_{I^*+1}-c+\sum_{i=1}^{I^*}b_i\right)+\sum_{i=I^*+2}^n(d_i-2)\beta_i\\
         =&\sum_{i=1}^{I^*}(d_i-2)(\beta_i-b_i)+
         (d_{I^*+1}-2)\left(\beta_{I^*+1}-c+\sum_{i=1}^{I^*}b_i\right)\\
         &+\sum_{i=I^*+2}^l(d_i-2)\beta_i+\sum_{i=l+1}^n(d_i-2)\beta_i.
         \end{split}
    \end{align}
In this case bounding the right-hand side from below is not as easy as the $\beta_i$ are linked together via the constraint $\sum_{i=1}^n\beta_i=\alpha$. Nevertheless, the last term is nonnegative for all $\beta\in X$, whereas all the terms in the first three terms of the sum are monotonically decreasing in $\beta_i$ and $\beta_{I^*+1}$, respectively. Consequently, we set $\beta_i=0$ for $i>l$ which monotonically increases the values of the remaining  $\beta_i$ and thus 
    \begin{align*}
      &\nabla F(x^*)^T(x-x^*)\\
      \geq&\sum_{i=1}^{I^*}(d_i-2)(\beta_i-b_i)+
         (d_{I^*+1}-2)\left(\beta_{I^*+1}-c+\sum_{i=1}^{I^*}b_i\right)+\sum_{i=I^*+2}^l(d_i-2)\beta_i\\
        =&\sum_{i=1}^{I^*}d_i(\beta_i-b_i)-2\sum_{i=1}^{I^*}\beta_i+2\sum_{i=1}^{I^*}b_i+d_{I^*+1}\left(\beta_{I^*+1}+\sum_{i=1}^{I^*}b_i-c\right)\\
        &-2\beta_{I^*+1}+2c-2\sum_{i=1}^{I^*}b_i
        +\sum_{i=I^*+2}^ld_i\beta_i-2\sum_{i=I^*+2}^l\beta_i\\
        =&\sum_{i=1}^{I^*}d_i(\beta_i-b_i)-2\sum_{i=1}^l\beta_i+d_{I^*+1}\left(\beta_{I^*+1}+\sum_{i=1}^{I^*}b_i-c\right)+2c+\sum_{i=I^*+2}^ld_i\beta_i\\
        =&\sum_{i=1}^{I^*}d_i(\beta_i-b_i)+2(c-\alpha)+d_{I^*+1}\left(\beta_{I^*+1}+\sum_{i=1}^{I^*}b_i-c\right)+\sum_{i=I^*+2}^ld_i\beta_i,
        \end{align*}
where we used the improved constraint $\sum_{i=1}^l\beta_i=\alpha$. Using the constraint again and rearranging terms gives
\begin{align}
\begin{split}
\label{variational_inequality_minimum_c}
        &\nabla F(x^*)^T(x-x^*)\\
        \geq&\sum_{i=1}^{I^*}d_i(\beta_i-b_i)+2(c-\alpha)+d_{I^*+1}\left(\alpha-\sum_{i=1}^{I^*}\beta_i-\sum_{i=I^*+2}^l\beta_{i}+\sum_{i=1}^{I^*}b_i-c\right)+\sum_{i=I^*+2}^ld_i\beta_i\\         
=&\sum_{i=1}^{I^*}d_i(\beta_i-b_i)+2(c-\alpha)+d_{I^*+1}(\alpha-c)+d_{I^*+1}\sum_{i=1}^{I^*}(b_i-\beta_i)
      +\sum_{i=I^*+2}^l(d_i-d_{I^*+1})\beta_i\\
         =&\sum_{i=1}^{I^*}\underbrace{(d_{I^*+1}-d_i)}_{\geq 0}\underbrace{(b_i-\beta_i)}_{\geq 0}+\underbrace{(c-\alpha)}_{\geq 0}\underbrace{(2-d_{I^*+1})}_{\geq 0}+\sum_{i=I^*+2}^l\underbrace{(d_i-d_{I^*+1})}_{\geq 0}\beta_i\geq 0
        \end{split}
    \end{align}
for all $x\in X$. Note that we used once more that $I^*+1\leq l$, so that $d_{I^*+1}\leq 2$. From \eqref{variational_inequality_minimum_c} we conclude that $x^*$ solves the variational inequality \eqref{variational_inequality} also in the case $\alpha^*=c$ and $x^*$ is thus the global minimum of $F$.
\end{proofof}

\begin{remark}
\label{rem:derive_optimal_choice}
Now we give a heuristic for the optimal choice $(\alpha^*,\beta^*)$ given in equation \eqref{optimal_choices_alpha_beta}. In a first step, we fix the amount of mass $\alpha$ that we want to transport and look for the optimal weights $\beta$ given $\alpha$. As $\beta_i=0$ for all $i=1,...,n$ if $\alpha=0$,  we will only consider $\alpha>0$. According to \eqref{total_bound_alpha_beta} we have the estimate
    \begin{align}
    \label{bound_alpha_>_0}
     \rho_F(\mu,\nu)\leq c+\sum_{i=1}^nb_i-2\alpha+\sum_{i=1}^n\beta_id_i. 
    \end{align}
As the Wasserstein distance scales with the transported distance, it is clearly optimal to transport mass as shortly as possible. Since the $d_i$ are ordered increasingly, this means that we prioritize lower indices over higher ones when assigning mass for transportation. To this end, let 
    \begin{align*}
    I_{\alpha}:=\max\left\{I\in \{0,...,n\}\mid \sum_{i=1}^Ib_i\leq \alpha\right\}, 
    \end{align*}
i.e. $I_{\alpha}$ denotes the index up to which we can  assign the maximal value to $\beta_i$, i.e. $\beta_i=b_i$, as there is still sufficient mass left assigned for transportation. The remaining mass $\alpha-\sum_{i=1}^{I_{\alpha}}b_i$ is then delegated to the next entry $I_{\alpha}+ 1$. All other entries are set to zero, so that this scheme yields the following weight vector $\beta^*=\beta^*(\alpha)$:
    \begin{align}
    \label{optimal_beta_i}
        \beta_i^*=\begin{cases}
         b_i,&i\leq I_{\alpha},\\
         \alpha-\sum_{i=1}^{I_{\alpha}}b_i, &i=I_{\alpha}+1,\\
         0,& \text{ else.}
        \end{cases}
    \end{align}
We note that distributing the mass in any other way by choosing different $\beta_i$ can not yield a better 
overall transportation cost $W_1(\tilde \mu_{\alpha},\nu_{\alpha,\beta})$ as we would potentially transport more mass to locations further away at the expense of nearer locations. However, equally efficient transport plans might be possible if there are points with the same distance to $x_0$ so that mass transportation is indifferent between those locations. \\
Now we are left to find the optimal choice of $\alpha$ which minimizes \eqref{bound_alpha_>_0}. So we define
    \begin{align}
    \begin{split}
    \label{definition_f_alpha}
       \tilde F(\alpha):=&F(\alpha,\beta^*)=c+\sum_{i=1}^nb_i-2\alpha+\sum_{i=1}^n\beta_i^*d_i\\
       =&c+\sum_{i=1}^nb_i-2\alpha+\sum_{i=1}^{I_{\alpha}}b_id_i+\left(\alpha-\sum_{i=1}^{I_{\alpha}}b_i\right)d_{I_{\alpha}+1}\\
       =&\underbrace{c+\sum_{i=1}^nb_i}_{=:K}+\sum_{i=1}^{I_{\alpha}}b_i(d_i-2)+(d_{I_{\alpha}+1}-2)\left(\alpha-\sum_{i=1}^{I_{\alpha}}b_i\right)
       \end{split}
       \end{align}
and want to minimize with respect to $\alpha$.\\
We claim that $\alpha$ is actually bounded by $\sum_{i=1}^l b_i$, so that 
	\begin{align}
	\label{correct_bounds_alpha_I_alpha}
	\alpha\leq\min\left\{c,\sum_{i=1}^lb_i\right\}\qquad \text{and}\qquad   I_{\alpha}=\max\left\{I\in \{0,...,l\}\mid \sum_{i=1}^Ib_i\leq \alpha\right\}.
	\end{align}
To this end, first consider the case that $\alpha> \sum_{i=1}^{l}b_i$. Then clearly $I_{\alpha}\geq l$. If $I_\alpha=l$, then we see
	\begin{align}
	\begin{split}
	\label{eq:alpha_geq_sum_can_not_be_optimal}
	\tilde F_l:=\tilde F(\alpha)=&K+\sum_{i=1}^{l}b_i(d_i-2)+\underbrace{(d_{l+1}-2)}_{>0}\underbrace{\left(\alpha-\sum_{i=1}^{l}b_i\right)}_{\geq 0}\\
	>&K+\sum_{i=1}^{l}b_i(d_i-2)=F\left(\sum_{i=1}^lb_i\right),
	\end{split}
	\end{align}
so that choosing $\alpha=\sum_{i=1}^lb_i$ leads to a smaller value. On the other hand, if $I_{\alpha}\geq l+1$, then due to the monotonicity of the $d_i$
	\begin{align*}
	\tilde F(\alpha)=&K+\sum_{i=1}^{l}b_i(d_i-2)+\sum_{i=l+1}^{I_{\alpha}}b_i(d_i-2)+(d_{I_{\alpha}+1}-2)\left(\alpha-\sum_{i=1}^{I_{\alpha}}b_i\right)\\
	\geq&K+\sum_{i=1}^{l}b_i(d_i-2)+\sum_{i=l+1}^{I_{\alpha}}b_i(d_i-2)+(d_{l+1}-2)\left(\alpha-\sum_{i=1}^{I_{\alpha}}b_i\right)\\
	=&K+\sum_{i=1}^{l}b_i(d_i-2)+ (d_{l+1}-2)\left(\alpha-\sum_{i=1}^{l}b_i\right) +\sum_{i=l+1}^{I_{\alpha}}b_i(d_i-2)-\sum_{i=l+1}^{I_{\alpha}}(d_{l+1}-2)b_i\\
	=&\tilde F_l+\sum_{i=l+1}^{I_{\alpha}}b_i\underbrace{(d_i-d_{l+1})}_{\geq 0}\geq \tilde F_l.
	\end{align*}
As we have seen in \eqref{eq:alpha_geq_sum_can_not_be_optimal} $\tilde F_l$ can not be minimal and therefore neither any $\alpha$ with $I_{\alpha}\geq l+1$. We conclude that indeed  \eqref{correct_bounds_alpha_I_alpha} holds. \\

\noindent
We come back to minimizing $\tilde F$ in \eqref{definition_f_alpha} and claim that $\tilde F$ is monotonically decreasing for $\alpha\in \left[0,\min\left\{c,\sum_{i=1}^lb_i\right\}\right]$ which will conclude this proof. Indeed, if $\tilde F$ is montonically decreasing, it takes its minimum at the right hand side of the interval and thus $\alpha^*=\min\left\{c,\sum_{i=1}^lb_i\right\}$.\\

\noindent
For the proof of monotonicity, we first note that without loss of generality we can assume $I_{\alpha}+1\leq l$, so that in the last term of \eqref{definition_f_alpha} we have $d_{I_{\alpha}+1}\leq 2$. Indeed, 
$I_{\alpha}+1>l$ implies $I_{\alpha}>l-1$, or rather $I_{\alpha}=l$ and thus $\alpha=\sum_{i=1}^lb_i$ with the third term vanishing completely in this case.\\
Now we turn to the monotonicity of $\tilde F$.  Let $\alpha_1\leq \alpha_2\leq \min\left\{c,\sum_{i=1}^lb_i\right\}$. If  $I_{\alpha_1}=I_{\alpha_2}=:I$, then
	\begin{align*}
	\tilde F(\alpha_1)-\tilde F(\alpha_2)=\underbrace{(d_{I+1}-2)}_{\leq 0}(\alpha_1-\alpha_2)\geq 0, 
	\end{align*}
so $\tilde F$ is indeed monotonically decreasing in this case. If  $I_{\alpha_1}<I_{\alpha_2}\leq l$, then 
	\begin{align*}
	&\tilde F(\alpha_1)-\tilde F(\alpha_2)\\
	=&K+\sum_{i=1}^{I_{\alpha_1}}b_i(d_i-2)+(d_{I_{\alpha_1}+1}-2)\left(\alpha_1-\sum_{i=1}^{I_{\alpha_1}}b_i\right)\\
	&-K-\sum_{i=1}^{I_{\alpha_2}}b_i(d_i-2)-(d_{I_{\alpha_2}+1}-2)\left(\alpha_2-\sum_{i=1}^{I_{\alpha_1}}b_i\right)\\
	=&-\sum_{i=I_{\alpha_1}+1}^{I_{\alpha_2}}b_id_i+(d_{I_{\alpha_1}+1}-2)\alpha_1-d_{I_{\alpha_1}+1}\sum_{i=1}^{I_{\alpha_1}}b_i-(d_{I_{\alpha_2}+1}-2)\alpha_2+d_{I_{\alpha_2}+1}\sum_{i=1}^{I_{\alpha_2}}b_i.
	\end{align*}
We rearrange and both add and substract auxiliary terms which leads to	
	\begin{align*}
	&\tilde F(\alpha_1)-\tilde F(\alpha_2)\\
	=&d_{I_{\alpha_2}+1}\sum_{i=1}^{I_{\alpha_2}}b_i-d_{I_{\alpha_1}+1}\sum_{i=1}^{I_{\alpha_2}}b_i+d_{I_{\alpha_1}+1}\sum_{i=1}^{I_{\alpha_2}}b_i
	-\sum_{i=I_{\alpha_1}+1}^{I_{\alpha_2}}b_id_i-d_{I_{\alpha_1}+1}\sum_{i=1}^{I_{\alpha_1}}b_i\\
	&+(d_{I_{\alpha_1}+1}-d_{I_{\alpha_2}+1})\alpha_1+d_{I_{\alpha_2}+1}(\alpha_1-\alpha_2)+2(\alpha_2-\alpha_1)\\
	=&(d_{I_{\alpha_2}+1}-d_{I_{\alpha_1}+1})\left(\sum_{i=1}^{I_{\alpha_2}}b_i-\alpha_1\right)+d_{I_{\alpha_1}+1}\sum_{i=I_{\alpha_1}+1}^{I_{\alpha_2}}b_i-\sum_{i=I_{\alpha_1}+1}^{I_{\alpha_2}}b_id_i\\
	&+(\alpha_2-\alpha_1)(2-d_{I_{\alpha_2}+1})\\
	=&(d_{I_{\alpha_2}+1}-d_{I_{\alpha_1}+1})\left(\sum_{i=1}^{I_{\alpha_1}}b_i-\alpha_1\right)+(d_{I_{\alpha_2}+1}-d_{I_{\alpha_1}+1})\sum_{i=I_{\alpha_1}+1}^{I_{\alpha_2}}b_i\\
	&+\sum_{i=I_{\alpha_1}+1}^{I_{\alpha_2}}b_i(d_{I_{\alpha_1}+1}-d_i)+(\alpha_2-\alpha_1)(2-d_{I_{\alpha_2}+1})\\
	=&(d_{I_{\alpha_2}+1}-d_{I_{\alpha_1}+1})\left(\sum_{i=1}^{I_{\alpha_1}}b_i-\alpha_1\right)+\sum_{i=I_{\alpha_1}+1}^{I_{\alpha_2}}b_i(d_{I_{\alpha_2}+1}-d_i)+(\alpha_2-\alpha_1)(2-d_{I_{\alpha_2}+1}).
	\end{align*}
Now using that $\sum_{i=1}^{I_{\alpha_1}+1}b_i-\alpha_1> 0$ by definition of $I_{\alpha_1}$, we conclude that $\tilde F$ is also monotonically decreasing in this case
	\begin{align*}
	&\tilde F(\alpha_1)-\tilde F(\alpha_2)\\
	=&\underbrace{(d_{I_{\alpha_2}+1}-d_{I_{\alpha_1}+1})}_{\geq 0}\underbrace{\left(\sum_{i=1}^{I_{\alpha_1}+1}b_i-\alpha_1\right)}_{>0}+\underbrace{\sum_{i=I_{\alpha_1}+2}^{I_{\alpha_2}}b_i(d_{I_{\alpha_2}+1}-d_i)}_{\geq 0}+\underbrace{(\alpha_2-\alpha_1)(2-d_{I_{\alpha_2}+1})}_{\geq 0}\\
	\geq& 0.
	\end{align*}

\noindent
It is important to note that this approach just provides a heuristic for the global minimum $(\alpha^*,\beta^*)$ and that a proof for the optimality is still necessary. In this remark we computed 
    \begin{align*}
        \min_{\alpha}\min_{\beta} F(\alpha,\beta)\qquad \text {instead of }\qquad \min_{\alpha,\beta}F(\alpha,\beta)
    \end{align*}
under the constraint $\sum_{i=1}^n \beta_i=\alpha$. In general, the second term is smaller than the first.
\end{remark}
\end{appendices}

\bibliography{bibliography}

\end{document}